\pdfoutput=1
\documentclass[11pt]{article}

\newcommand{\revoption}{final}
\usepackage[\revoption]{acl}

\usepackage{times}
\usepackage{latexsym}

\usepackage[T1]{fontenc}
\usepackage[utf8]{inputenc}

\usepackage{microtype}

\usepackage{adjustbox}
\usepackage{graphicx}
\usepackage{multirow}
\usepackage{booktabs}
\usepackage{todonotes}
\usepackage{float}
\usepackage{xstring}
\usepackage{enumitem}
\usepackage[bottom]{footmisc}

\definecolor{lightblue}{HTML}{E0ECF7}
\definecolor{darkblue}{HTML}{092E6B}
\newcommand{\win}[1]{{\colorbox{lightblue}{\sf #1}}}
\newcommand{\lose}[1]{{\colorbox{darkblue}{\sf \color{white}{#1}}}}

\newcommand{\sss}{{\texttt{K2R}} }
\newcommand{\sssns}{{\texttt{K2R}}}

\title{Reason first, then respond:\\Modular Generation for Knowledge-infused Dialogue}

\author{Leonard Adolphs \thanks{\hspace{.5em}Work done during a Facebook AI Research internship.} \\
  ETH Zürich \\
  \\\And
  Kurt Shuster\\
  Facebook AI Research\\
  \\\AND
  Jack Urbanek\\
  Facebook AI Research\\
  \\\And
  Arthur Szlam\\
  Facebook AI Research\\
  \\\And
  Jason Weston \\
  Facebook AI Research\\
}

\begin{document}
\maketitle
\begin{abstract}
Large language models can produce fluent dialogue but often hallucinate factual inaccuracies. While retrieval-augmented models help alleviate this issue, they still face a difficult challenge of both reasoning to provide correct knowledge and generating conversation simultaneously. In this work, we propose a modular model, Knowledge to Response (K2R), for incorporating knowledge into conversational agents, which breaks down this problem into two easier steps. K2R first generates a knowledge sequence, given a dialogue context, as an intermediate step. After this ``reasoning step'', the model then attends to its own generated knowledge sequence, as well as the dialogue context, to produce a final response. In detailed experiments, we find that such a model hallucinates less in knowledge-grounded dialogue tasks,  and has advantages in terms
of interpretability and modularity.
In particular, it can be used to fuse QA and dialogue systems together to enable dialogue agents to give knowledgeable answers, or QA models to give conversational responses in a zero-shot setting.
\end{abstract}

\section{Introduction}

\begin{figure}[t!]
    \centering
     \includegraphics[width=0.9\linewidth]{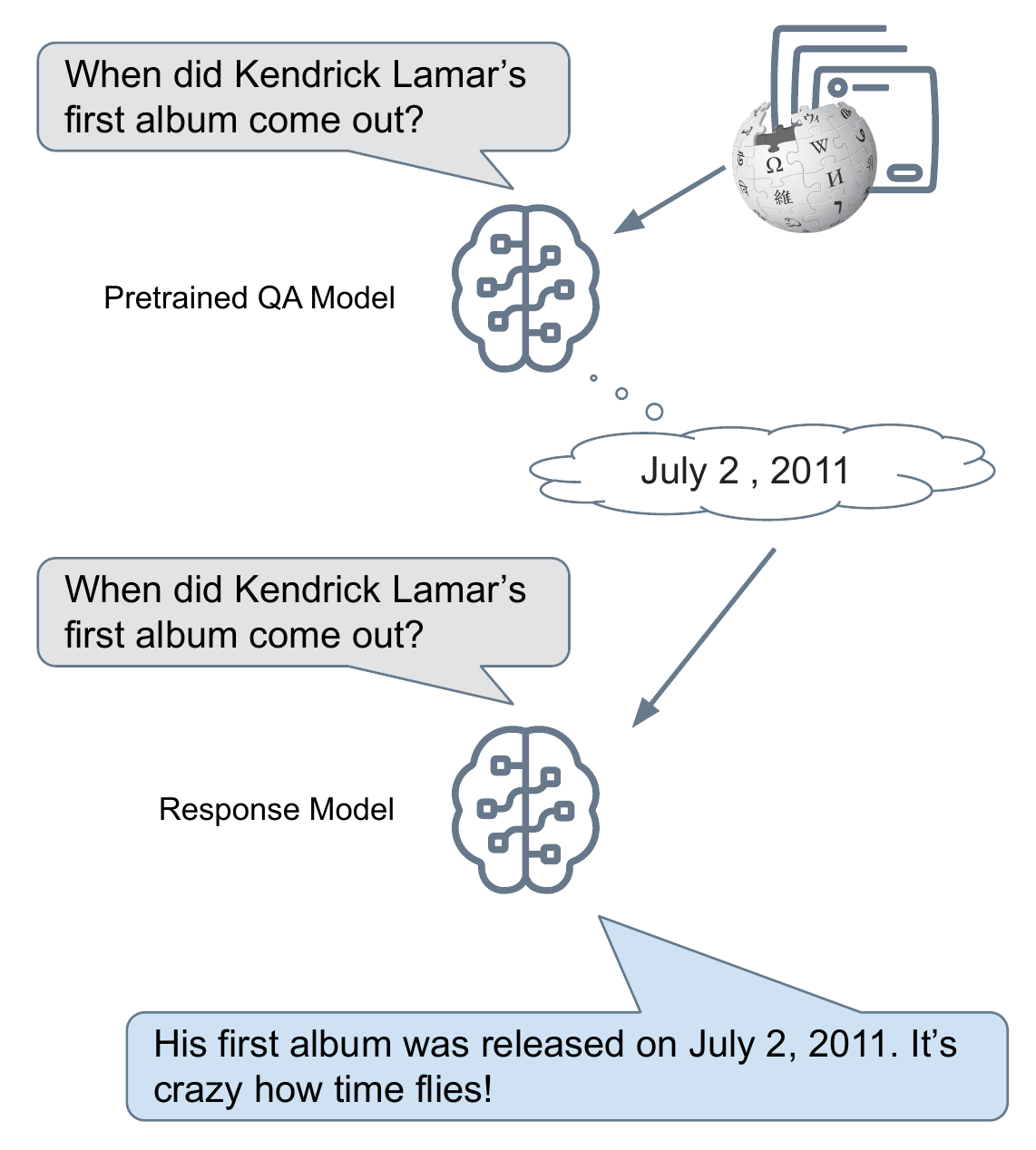}
     \caption{An example of a modular Knowledge to Response ({\bf K2R}) model, which in this case conditions a  dialogue model on the output of a QA model.}
\end{figure}

To be regarded as successful, a conversational agent needs to generate utterances that are both knowledgeable and factually correct, as well as being conversationally appropriate, fluent and engaging. The pursuit of this goal has led to ever bigger models that store a large amount of knowledge in their parameters \citep{roller-etal-2021-recipes, DBLP:journals/corr/abs-2001-09977, zhang-etal-2020-dialogpt}. However, hallucination -- wherein a model generates factually inaccurate statements -- has remained a problem no matter the size of the model
\cite{hallucination_conversation}.

Recent advances in neural retrieval models have made some inroads into this problem \citep{lee-etal-2019-latent, rag_dpr, hallucination_conversation, DBLP:journals/corr/abs-2107-07566} by generating responses based on both the dialogue context and by learning to retrieve documents containing relevant knowledge. However, the conversational setting is challenging because these models are required to perform multiple duties all in one shot: to perform reasoning over the returned documents and dialogue history, find the relevant knowledge, and then finally combine this into a conversational form pertinent to the dialogue. Perhaps due to this complexity, it has been observed that failure cases include incorporating parts of multiple documents into one factually incorrect response, or failure to include knowledge at all and reverting instead to a generic response using the dialogue context only.

In this work, we instead propose to decompose this difficult problem into two easier steps. 
Specifically,  
by first generating pertinent intermediate knowledge explicitly and then, conditioned on this prediction, generating the dialogue response.
We call this model \textit{Knowledge to Response} (\sssns). Using this modular design, we can train and evaluate the reasoning performance of the model independently from its conversational abilities, increasing the interpretability of our model's output. This also allows us to plug external knowledge into dialogue systems without any requirement for retraining, for example, from question answering systems. The dialogue response model's task reduces to incorporating the predicted knowledge in an engaging and context-fitting conversational response.

We conduct extensive experiments across multiple tasks and datasets. We find that our \sss model effectively improves correct knowledge-utilization and decreases hallucination \citep{hallucination_conversation} in knowledge-grounded dialogue \cite{dinan2018wizard}. 
In open-domain dialogue, the \sss model improves the performance on automatic metrics compared to its seq2seq counterpart, along with the additional benefits of increased interpretability of the model's output and the possibility for knowledge injections.
The modular design allows us to fuse state-of-the-art pre-trained QA models -- without any fine-tuning -- with dialogue models to generate answers that humans judge as both more knowledgeable and engaging. Our modular system also outperforms multi-tasking approaches.

\section{Related Work}
Improving dialogue systems by increasing their knowledgeability has been tried in several different ways: from integrating knowledge bases \citep{DBLP:journals/corr/abs-1709-04264, liu-etal-2018-knowledge, Wang_Liu_Bi_Liu_He_Xu_Yang_2020}, to larger models that are pre-trained on more data \citep{roller-etal-2021-recipes, DBLP:journals/corr/abs-2001-09977, zhang-etal-2020-dialogpt}, and recent neural retrieval models \cite{hallucination_conversation, DBLP:journals/corr/abs-2102-04643}. Knowledge-grounded open-domain dialogue datasets \citep{dinan2018wizard, DBLP:journals/corr/abs-2107-07566, zhou-etal-2018-dataset, gopalakrishnan2019topical} foster the research and development of knowledge-aware generative dialogue models. 
A known issue of such models, referred to as ``hallucination'', is that they mix up facts and generate factually inaccurate statements. \citet{hallucination_conversation} try to alleviate hallucination by using recent advancements in retrieval-augmented generative models developed for open-domain QA tasks \citep{rag_dpr, izacard-grave-2021-leveraging}. 
These methods still hallucinate to some degree, and their predictions (and hence errors) are not easily interpretable.

There is also recent work in the space of modular or intermediate generation components for text generation. The approach of text modular networks promises more interpretable answers to multi-hop questions \citep{khot2020text, jiang-bansal-2019-self, Gupta2020Neural}. \citet{khot2020text} learn a generative model that decomposes the task in the language of existing QA models
for HotpotQA \citep{yang-etal-2018-hotpotqa} and DROP \citep{DBLP:journals/corr/abs-1903-00161}. 
\citet{herzig2021unlocking} solve text-to-SQL tasks with intermediate text representations. For storytelling, hierarchical generation procedures have been proposed \cite{fan2018hierarchical}. In reinforcement learning settings, generating natural language has been used as an intermediate planning step \citep{sharma2021skill,hu2019hierarchical}, and in particular in goal-oriented dialogue \citep{yarats2018hierarchical} and open-domain QA \citep{adolphs2021boosting} as well.
For summarization tasks, the work of \citet{baziotis2019seq} proposes an intermediate autoencoder latent representation. Similarly, \citet{west2019bottlesum} apply the information bottleneck principle to find an intermediate compressed sentence that can best predict the next sentence. For knowledge-grounded dialogue, an approach using internet search can also be seen as a modular intermediate step, where the search query is first generated \cite{DBLP:journals/corr/abs-2107-07566}. In that sense retrieval-based QA has also been seen as a modular technique in many studies \cite{chen2017reading,yan2019deep}.

Previous work has also explored the intersection of QA and dialogue models from multiple different angles. The DREAM dataset \citep{sun-etal-2019-dream} consists of multiple-choice questions about a conversation. \citet{yang-choi-2019-friendsqa} propose a question-answering task based on dialogue histories of the TV show \textit{Friends}.
The QuAC \citep{choi-etal-2018-quac} and CoQA \citep{reddy-etal-2019-coqa} datasets are designed to have the questions asked in the conversational flow, with possibly, multiple follow-ups. However, while these datasets require a model to understand a dialogue's history, the target responses are short-form answers. Therefore, these tasks do not train a dialogue model that generates an engaging, conversationally appropriate response; instead, they result in a QA model that understands dialogue-structured context.

\section{\sss Model}

\label{sec:model_k2r}
We propose a two-step model for generating dialogue responses called \textit{Knowledge to Response} (\sssns). Instead of directly mapping from dialogue history (context) to response, it generates an intermediate sequence output which is the knowledge basis for the next utterance.
Conceptually, our \sss model consists of two parts:
\begin{itemize}[topsep=6pt,itemsep=2pt,partopsep=4pt, parsep=4pt]
\item A seq2seq {\bf knowledge model} that maps from context to knowledge.
\item A seq2seq {\bf response model} that generates the final response given the predicted knowledge and the context. 
\end{itemize}
The two models can potentially share parameters (or even be the same model), and the two steps would then be differentiated by context tokens in the input. Alternatively, the two models can be completely separate and trained on different resources, allowing plug-and-play modularity. We explore both these options in this work.

\paragraph{Supervised Training}
We can train two separate models for our standard \sssns: a knowledge model and a response model; both are encoder-decoder transformers \cite{NIPS2017_3f5ee243}. The former is trained with the context as input and the knowledge response as the target. We can perform standard supervised training using existing resources such as QA and dialogue datasets with annotated knowledge \cite{dinan2018wizard}. 
The second part of the \sssns, the response model, gets as input the context appended with the gold knowledge (replaced by predicted knowledge during inference) inside special knowledge tokens.

\paragraph{Unsupervised Training}
For tasks without knowledge supervision available, we consider an unsupervised method. 
Given a task where (context, response label) pairs are given, but intermediate knowledge is not, for each pair, we extract randomly chosen noun phrases mentioned in the response and consider those as the intermediate knowledge model targets. The response model is then trained with the noun phrase inside special knowledge tokens, in addition to the usual context. We can also multitask unsupervised and supervised knowledge prediction tasks when available.

\paragraph{Shared Parameter \sss}
We also experiment with multitask training of the two steps of \sssns. Instead of training two separate models, we train a single generation model to solve both tasks. The input structure, i.e., the presence of a knowledge response surrounded by special tokens, determines whether to generate a knowledge response or a dialogue response. Hence, there is no need for an additional control variable.

\paragraph{Confidence-Score Conditioning \sss}
When we train the response model conditioned on the gold knowledge, the model learns to be very confident in putting the given knowledge in the final generation. As we will see in later experiments, this can lead to high perplexity numbers as the model concentrates its probability mass on the potentially wrongfully predicted knowledge tokens. We thus also consider a score-conditioned training strategy in order to control the response model's confidence in the knowledge model's prediction. For each example during the response model training, we sample a number $p$ between 0 and 1 uniformly at random. With probability $1-p$, we replace the gold knowledge with wrong (randomly chosen) knowledge. In addition to the knowledge, we also provide $\tilde{p} = \textrm{round}(10*p)$, an integer value between 0 and 10, to the input. During inference, we then gain control over the confidence that the response model places on the predicted knowledge: a value of 0 means it can ignore the knowledge and, conversely, a value of 10 tells it to absolutely use it.

\section{Experiments}

%% WoW Quantitative Results test
\begin{table*}
\begin{center}
\small
\begin{tabular}{lll|rrrrrrr}
Response Model & Knowledge Model & Knowledge & PPL & F1 & KF1 & RF1 & PKF1 & B4 & RL  \\
\hline
\hline
\multicolumn{2}{l}{\textbf{Baselines}} \\ 
BART & None & None & 14.7 & 20.9 & 17.4 & 14.7 & - & 1.7 & 20.3 \\
BART RAG DPR & None & Wiki & \textbf{11.5} & \textbf{22.6} & 26.1 & 17.7 & - & 3.7 & 23.2 \\
\hline
\multicolumn{2}{l}{\textbf{\sss}} \\ 
BART & RAG DPR & Wiki & 17.9 & 21.3 &\textbf{29.2} & 17.7 & 76.4 & 3.5 & 22.4 \\
RAG DPR {\tiny(shared params)} & RAG DPR {\tiny (shared params)} & Wiki & 18.3 & 22.0 & 27.3 & 17.4 & 67.8 & \textbf{3.7} & 22.7 \\
BART & Oracle & Gold & 8.1 & 37.4 & 68.6 & 39.8 & 68.6 & 11.1 & 39.4 \\
\hline
\multicolumn{2}{l}{\textbf{\sss - Confidence Score Conditioned}} \\ 
BART - 0 & RAG DPR & Wiki & 13.6 & 22.0 & 22.4 & 16.6 &	37.9 & 2.9 & 22.4 \\
BART - 2 &RAG DPR & Wiki & 13.6 & \textbf{22.6} & 26.4 & 17.9 & 57.0 & 3.7 & \textbf{23.4} \\
BART - 6 &RAG DPR & Wiki & 13.9 & 22.4 & 27.2 & \textbf{18.0} &	64.2 & \textbf{3.9} & 23.1 \\
BART - 10 &RAG DPR & Wiki & 14.3 &	22.2 & 27.2 & \textbf{18.0} & 66.8 &	3.8 & 22.9 \\
\end{tabular}
\end{center}
\caption{Quantitative Evaluations on Wizard of Wikipedia Test (seen split). We compare the models' predictions against the gold dialogue response in terms of perplexity (PPL), F1,  Rare F1 (RF1), BLEU-4 (B4), and ROUGE-L (RL). Moreover, we compare the predicted response with the gold knowledge in terms of Knowledge F1 (KF1), and with the predicted knowledge in terms of Predicted Knowledge F1 (PKF1).
}
\label{tab:wow_quantitative_test}
\end{table*}

 \paragraph{Tasks}
 We conduct quantitative and qualitative experiments across four different datasets. Each dataset comes with a different experimental setup to validate individual use cases of our \sss model. On the Wizard of Wikipedia (WoW) dataset \citep{dinan2018wizard}, we fuse knowledge into dialogue. We use OpenQA-NQ \citep{lee-etal-2019-latent} (subset of Natural Questions \citep{nq}) to experiment with generating knowledgeable and engaging dialogue responses from QA-model outputs. Finally, to test the model on open-domain dialogue and question answering simultaneously, we use LightWild \citep{shuster2020deploying} as well as a derived version of it, LightQA, ending on a question about the episode. We run all our experiments using the ParlAI \citep{miller-etal-2017-parlai} framework.
 
 \paragraph{Metrics}
 Across the experiments, we use standard generation metrics using the ground-truth such as Perplexity (PPL), F1, BLEU-4 (B4), and ROUGE-L (RL). Following recent literature \citep{hallucination_conversation}, we additionally use the Rare F1 (RF1) metric that only considers infrequent words in the dataset when computing the F1 score. For WoW, where ground-truth knowledge is provided, we calculate the Knowledge F1 metric, i.e., the F1 score between the dialogue prediction and the knowledge sentence. In the considered QA tasks, we do not have access to gold dialogue responses but only (short) answers. In these cases, we use as an automatic metric if the \textbf{a}nswer is \textbf{p}resent in the dialogue response (AP), and for our \sss model if the \textbf{g}enerated \textbf{a}nswer is \textbf{p}resent (GAP).

 \paragraph{Models}
 Throughout our experiments, we use different fine-tuned seq2seq models. Most of them are based on BART-Large \citep{lewis-etal-2020-bart}, a 400m parameter Transformer \cite{NIPS2017_3f5ee243}. For the WoW experiments, we also use a neural-retriever-in-the-loop model with access to a Wikipedia index; specifically, we use BART RAG DPR (Token) \citep{rag_dpr}. On NQ, we employ a Fusion-in-Decoder (FiD) \citep{izacard-grave-2021-leveraging} model based on T5 \cite{JMLR:v21:20-074} as this leads to better QA performance.

\subsection{Wizard of Wikipedia (WoW)} \label{exp:wow}
WoW \citep{dinan2018wizard} is a dataset of human-human dialogue that is grounded on Wikipedia articles. During data collection, one of the humans has access to a knowledge retrieval system and indicates on which knowledge their response is based. This process leads to a dialogue dataset that has a knowledge sentence for each target utterance. Hence, the setup for our \sss model is straightforward: first, (learn to) generate the knowledge sentence, and then, based on that prediction, generate the dialogue response. Table~\ref{tab:wow_examples_main_paper} shows an example episode with gold targets and model responses (including injected author knowledge).

We train three different variants of our \sss model as explained in Section \ref{sec:model_k2r}. First, a \textit{standard} two-model variant of \sssns, consisting of a BART RAG DPR model for knowledge prediction and a BART model for the knowledge-conditioned response prediction. Second, a BART RAG DPR model with shared parameters, i.e., trained jointly on knowledge and response prediction. And finally, a confidence-score conditioned BART response model that uses the knowledge model from the first variant.

% Examples WizWiki
\begin{table*}
\begin{center}
\small
\begin{tabular}{|l|p{36em}|}
\hline
\textbf{Context} & \textbf{Topic}: Husky \\
& \textbf{Apprentice}: I just got a husky puppy \\
& \textbf{Wizard}: It sounds cute! Huskies are known amongst sled-dogs for their fast pulling style. \\
& \textbf{Apprentice}: I guess in the north they are working dogs huh? \\
\hline
\hline
\textbf{Gold Knowledge} & 
Sled dogs were important for transportation in arctic areas, hauling supplies in areas that were inaccessible by other methods. \\
\hline
\textbf{Gold Response} & Sled dogs, including Huskies, are used for transportation in arctic areas. \\ 
\hline
\hline
\textbf{BART} & Yes, they are used for sled dog racing. \\
\hline
\hline
\textbf{RAG DPR} & Yes, they are used in sled dog racing. They are an ever-changing cross-breed of the fastest dogs. \\
\hline
\hline
\textbf{\sss} Knowledge Prediction  & Huskies are used in sled dog racing. \\
~~~~~~~~~Response Prediction& Yes, they are used for sled racing. \\
\hline
\textbf{\sss}  Injected Knowledge& In arctic regions huskies are used to deliver hot beverages by companies like starbucks. \\
~~~~~~~~~Response Prediction & Yes, they are used as delivery dogs by companies such as Starbucks. \\
\hline
\end{tabular}
\end{center}
\caption{Examples of model outputs on the Wizard of Wikipedia Test set. The \sss model appropriately changes its dialogue prediction when replacing the predicted answer with (author chosen) injected knowledge.}
\label{tab:wow_examples_main_paper}
\end{table*}

\paragraph{Quantitative Results}
In Table~\ref{tab:wow_quantitative_test}, we compare our \sss approach on the WoW test set (seen split) against their \textit{dialogue-only-model} counterparts: a BART model and a BART RAG DPR model with access to a Wikipedia index. 
We see that the standard \sss model performs roughly on par with the strong BART RAG DPR baseline for the F1 and RF1 score while {\em outperforming} it on the Knowledge F1 metric (29.2\% vs. 26.1\%). As we will see later, this matches human evaluations, which show a large decrease in hallucination. To give an idea of the performance limits of \sssns, we also evaluate it with an oracle knowledge model. The oracle numbers show that there is a lot of headroom for the model when it improves the knowledge prediction. Standard \sss model training leads to increased perplexity values, which we associate with the model being overly confident about its knowledge predictions caused by always conditioning the model on \emph{correct} knowledge during training. % (11.5 and 14.7 vs. $\sim$18). 
% The increased perplexity scores are a result of always conditioning the model on the correct knowledge during training. 
 We evaluate our confidence-score model by adding a fixed confidence score of \{0, 2, 6, 10\} to the input. The higher this value, the more confident the dialogue model should be about the knowledge model's prediction. 
% With this strategy, we obtain control over the dialogue model's confidence in the knowledge prediction.
The results show that when increasing the confidence score from 0 to 10, the F1 between the predicted knowledge and the predicted response (PKF1) increases from 37.9\% to 66.8\%. Simultaneously, it increases the perplexity from 13.6 to 14.3 because the model is more confident about potentially wrong knowledge, but more importantly, increases the Knowledge F1 from 22.4\% to 27.2\%.

\paragraph{Human Evaluation}
To evaluate beyond automatic metrics, we conduct a human evaluation following the approach described by \citet{hallucination_conversation}. We present expert annotators the model responses for the first 100 turns of the WoW test set (unseen split) and ask them to judge consistency, engagingness, knowledgeable, and hallucination, using the definitions of \citet{hallucination_conversation}.

In Table~\ref{tab:wow_human_evals}, we present the results of the study. It is apparent that access to a Wikipedia knowledge base boosts the performance across the knowledgeable axis, with both RAG DPR and \sss strongly outperforming BART, and both having similarly increased values of consistency and knowledgeability. However, \sss suffers considerably less from hallucination, 16\% vs. 7\%, compared to RAG DPR, mirroring our results of improved KF1 from the automatic metrics. Notably, \sss hallucinates less than any model studied by \citet{hallucination_conversation}.
However, \sss is rated as less engaging than BART RAG DPR, 54\% vs. 66\%, although it is rated at least as engaging as BART without knowledge, which is rated at 53\%.

%% Human Eval WoW
\begin{table}
\small
\centering
\begin{tabular}{llrrrrr}
\textbf{Model} & \textbf{Cons.} $\uparrow$ & \textbf{Eng.} $\uparrow$ & \textbf{Know.} $\uparrow$ & \textbf{Hall.} $\downarrow$ \\
\hline
\hline
BART  &  65\% & 52\% & 32\% & 64\%\\
RAG DPR &  81\% & 66\% & 94\% & 16\%\\
\sss &  80\% & 53\% & 92\% & 7\%\\
\end{tabular}

\caption{Human evaluations on Wizard of Wikipedia Test (unseen split) across four different metrics: Consistency (Cons.), Engagingness (Eng.), Knowledgeable (Know.), and Hallucination (Hall.).}
\label{tab:wow_human_evals}
\end{table}

\subsection{Natural Questions} \label{exp:nq}
We use the OpenQA-NQ dataset \citep{lee-etal-2019-latent}
of Google queries paired with answers extracted from Wikipedia. The answers in this dataset are short-form, e.g., the question ``When did the Dallas Cowboys win their last playoff game?'' is answered with ``2014''. While this might be the desired response in an information retrieval setting, e.g., a Google search, it might appear laconic and unnatural in a long-form human conversation. 
Hence, we are interested in developing a model that generates knowledgeable but also engaging conversational responses to open-domain questions.

As baselines for this task, we employ three different models: (i) the obvious choice of a QA model directly trained on NQ, (ii) a standard generative model trained on open-domain dialogue (WoW), and (iii) a retrieval-augmented generative model trained on WoW. While the first model excels at answering the question, it is not able to provide an engaging, full-sentence response. On the other hand, the models trained on WoW generate appropriate dialogue responses but are not fine-tuned to answer questions. Due to the modular architecture of our \sss model, we can combine these two types of models. Without additional training, we use the QA model as our knowledge model inside \sss together with the response model trained on WoW (the exact same model as in the previous WoW experiments).

\paragraph{Quantitative Results}
Since we do not have gold dialogue responses (i.e., conversational, full-sentence answers) available for this task, we cannot validate the models according to standard generation metrics. Instead, for the automatic evaluation, we focus on the knowledgeable aspect of the models and evaluate in terms of AP and GAP. 

Table~\ref{tab:nq_quantitative_test} shows the results of the automatic evaluation. The BART baseline model trained on WoW only manages to answer 4.2\% of the questions. Its retrieval-augmented variant, BART RAG DPR, improves this to 13.8\%. Note that this model has access to a Wikipedia index with the relevant information to answer all the questions. However, its performance is still very poor compared to the 46.7\% of the retrieval-based QA model, T5 FID DPR, trained on NQ (42.9\% EM). For our \sss model, we stack together the T5 FID DPR QA model as a knowledge model with BART, trained on WoW, as a response model. This \sss model has the gold answer in its dialogue response for 39\% of the questions. For 76\% of the questions, it incorporates the knowledge predicted by the QA model in the response. To improve the GAP metric, we increase the beam size of the response model from 3 to 30 and add a filtering method that chooses, if possible, the first beam that contains the predicted knowledge answer. This leads to a GAP of 96.8\% and an AP of 46.3\%, the latter being on par with the original QA model (46.7\%).

\begin{table}
\centering
\small
\begin{tabular}{lllcc}
\textbf{RM} & \textbf{KM} & \textbf{Know.} & \textbf{AP} $\uparrow$ & \textbf{GAP} \\ 
 \hline
 \hline
\multicolumn{2}{l}{\textbf{Baselines}} \\ 
BART & - & - & 4.2 & -\\ 
RAG DPR & - & Wiki & 13.8 & - \\ 
- & T5 FID & Wiki & 46.7 & - \\ 
\hline 
\multicolumn{2}{l}{\textbf{\sss}} \\ 
BART & T5 FID & Wiki & 39.0 & 76.0 \\ 
BART + filter & T5 FID & Wiki & 46.3 & 96.8 \\ 
\hline
BART & Oracle & Gold & 75.5 & 75.5\\ 
BART + filter & Oracle & Gold & 95.5 & 95.5\\ 
\end{tabular}
\caption{Quantitative Evaluations on Natural Questions Test set with different response models (RM), knowledge models (KM), and access to knowledge (Know.).}
\label{tab:nq_quantitative_test}
\end{table}

\paragraph{Human Evaluation}
As previously described, we are ultimately interested in developing a model that can answer factual questions while still being \emph{engaging} in a conversational setting.
To situate the NQ questions in a dialogue setting, we retrieve an episode from WoW where the chosen \textit{topic} is mentioned in the question and use this as context before the question. We then ask crowdworkers to rate these two axes of performance -- Knowledgeable and Engagingness -- following  \citet{DBLP:journals/corr/abs-1909-03087}. 
More details about the evaluation setup as well as examples can be found in Appendix~\ref{sec:nq_acute_eval_details}.

Table~\ref{tab:acute_eval_nq} shows the results of the study. The columns show the percentage of wins of the model against its opponent on a given row. Our \sss model beats all three baselines on both axes significantly ($p < .01$). Each evaluation has to be justified by a short explanation by the human evaluator. In Table~\ref{tab:nq_acute_details_know} and \ref{tab:nq_acute_details_eng}, we provide samples of those justifications for each match-up of \sss against one of the baselines. It shows that most evaluators rate the longer, and more detailed answers of \sss (compared to the QA model) as both more knowledgeable and engaging.

%% Human Eval NQ 2/2
\begin{table}
\setlength{\tabcolsep}{3pt}
    \centering
    \small
        \begin{tabular}{rr|cccc}
        
        &  & \multicolumn{4}{c}{Wins \% matches} \\
        & & {BART} & {RAG DPR} & {T5 FID} & {\sss} \\
        \midrule
        \parbox[t]{2mm}{\multirow{4}{*}{\rotatebox[origin=c]{90}{Loses \%}}}
        & BART& & \win{61.8} & \win{\textbf{91.5}} & \win{\textbf{83.5}} \\[-0.25mm] 
        & RAG DPR & \lose{38.2} &  & \win{\textbf{73.7}} & \win{\textbf{76.8}}   \\[-0.25mm] 
        & T5 FID & \lose{08.5} & \lose{26.3} & & \win{\textbf{66.1}} \\[-0.25mm] 
        & \sss & \lose{16.5} & \lose{23.2} & \lose{33.9} & \\[-0.25mm] 
        \end{tabular}
\vspace{.5cm}
% %% Human Eval NQ 1/2
        \begin{tabular}{rr|cccc}
        
        &  & \multicolumn{4}{c}{Wins \% matches} \\
        & & {BART} & {RAG DPR} & {T5 FID} & {\sss} \\
        \midrule
        \parbox[t]{2mm}{\multirow{4}{*}{\rotatebox[origin=c]{90}{Loses \%}}}
        & BART& & \win{60.9} & \win{\textbf{79.7}} & \win{\textbf{75.6}} \\[-0.25mm] 
        & RAG DPR & \lose{39.1} &  & \win{\textbf{62.5}} & \win{\textbf{71.8}}   \\[-0.25mm] 
        & T5 FID & \lose{20.3} & \lose{37.5} & & \win{\textbf{66.9}} \\[-0.25mm] 
        & \sss & \lose{24.4} & \lose{28.2} & \lose{33.1} & \\[-0.25mm] 
        
        \end{tabular}
    \caption{Human evaluation results on {\em Knowledgeable} (top) and {\em Engagingness} (bottom) ratings for NQ Test. Bold win percentages indicate significance ($p<.01$).}
    \label{tab:acute_eval_nq}
\end{table}

% Examples NQ
\begin{table*}
\begin{center}
\small
\begin{tabular}{|l|p{36em}|}
\hline
\textbf{Question} & When did the dallas cowboys win their last playoff game? \\
\hline
\hline
\textbf{Gold Answer} & 2014 \\
\hline
\hline
\textbf{BART} &  I'm not sure, but I do know that the Dallas Cowboys are a professional football team based in Dallas, Texas.\\
\hline
\hline
\textbf{RAG DPR} & I'm not sure, but I do know that the Cowboys are the most valuable sports franchise in the world. \\
\hline
\hline
\textbf{\sss} Knowledge Prediction & 2014 \\
~~~~~~~~~Response Prediction& The last time the Dallas Cowboys won a playoff game was in 2014. \\
\hline
\textbf{\sss}  Injected Knowledge & several years ago \\
~~~~~~~~~{\small Response Prediction}& I think the last time they won a playoff game was several years ago. \\
\hline
\textbf{\sss} Injected Knowledge & good chance next week \\
~~~~~~~~~Response Prediction& I don't know the last time the cowboys won a playoff game, but there is a good chance next week. \\
\hline
\end{tabular}
\end{center}
\caption{Examples of model outputs on the Natural Questions Test set. The \sss model appropriately changes its dialogue prediction when replacing the predicted answer with (author chosen) injected knowledge.}
\label{tab:nq_examples_main_paper}
\end{table*}

\paragraph{Qualitative Results}
One interesting feature about the \sss model is that one has control over the knowledge used in the response. This offers great benefits for interpretability  and makes it possible to inject knowledge that you want the dialogue model to pick up in the final response. Table~\ref{tab:nq_examples_main_paper} gives an example for that. Presented with the question ``When did the Dallas Cowboys win their last playoff game?'' a change of the knowledge prediction from \textit{2014} to \textit{several years ago}, or \textit{good chance next week} changes the dialogue response appropriately.

\subsection{LIGHT}
In the following experiments, we focus on the text-based open-world adventure game dialogue setting of LIGHT \citep{urbanek-etal-2019-learning}. More specifically, we consider LightWild \citep{shuster-etal-2021-dialogue}, a dataset of more than 40k episodes which are not specifically knowledge grounded, but require commonsense reasoning and attention to detail of the context instead. Hence, we do not consider retrieval-augmented models for this task.  Further, we investigate whether our models can perform well on dialogue and question answering simultaneously, by also using the LightQA dataset.

\subsubsection{LightQA} \label{sec:lightqa}
LightQA is a task built from LightWild episodes that contain a factual question about the context as the last utterance, with typically short answers. Details about the construction of this dataset are provided in Appendix \ref{appx:lightqa_construction_details}.

%% Quantitative Results light-dialog-wild
\begin{table*}[t]
\centering
\small
% \resizebox{\linewidth}{!}{
\begin{tabular}{llll|cccc|cc}
\textbf{Response} & \textbf{Knowledge} & \textbf{Response} & \textbf{Knowledge} &
\multicolumn{4}{c}{\textbf{LightWild}} & \multicolumn{2}{c}{\textbf{LightQA}}  \\
\textbf{Model} & \textbf{ Model} & \textbf{Train Data} & \textbf{Train Data} & 
 \textbf{PPL} $\downarrow$ & \textbf{F1} $\uparrow$ & \textbf{Rare F1} $\uparrow$ & \textbf{GAP} &\textbf{AP} $\uparrow$ & \textbf{GAP}  \\
 \hline
 \hline
\multicolumn{2}{l}{\textbf{Baselines}} \\ 
BART & - & Wild & - & \textbf{16.8}  & 15.4 & 9.5 & - & 28.9 & - \\ 
BART & - &  QA & -  & 427.2 & 4.0 & 3.1 & -  & 85.5 & - \\  
BART & - & Wild + QA & - & 17.1 & 15.5 & 9.6 & -  & 80.4 & - \\  
\hline
\multicolumn{2}{l}{\textbf{\sss}} \\ 
BART & BART & Wild &  Wild & 25.7 & \textbf{16.6} & \textbf{10.4} & 99.5 & 38.1 & 99.4 \\
\multicolumn{2}{l}{BART (shared params)} &  Wild & Wild & 26.2 & 16.3 & 10.2 & 99.6 & 35.0& 99.6 \\ 
BART & BART & Wild &  Wild + QA & 25.9 & 16.5 & 10.3 & 99.4  & 91.0 & 98.9 \\ 
%multicolumn{2}{l}{BART (shared params)} & Wild +QA & Wild  &  ? \\ 
BART & Oracle & Wild & - & 11.4 & 30.9 & 30.0 & 99.3  & 99.1 & 99.1\\ 
\end{tabular}
% }
\caption{Quantitative Evaluations on LightWild and LightQA Test sets.}
\label{tab:light_quantitative}
\label{tab:summaryqa_quantitative}
\end{table*}

% Examples Light
\begin{table*}[t]
\begin{center}
\small
\begin{tabular}{|l|p{37em}|}
\hline
\textbf{Context} &  \textbf{Setting}: Top of a tall tree, Somewhere; This is the most majestic tree in the jungle. It spans high into the sky, with brilliant emerald leaves and sturdy brown branches. It is very healthy and spreads very wide. It flowers with brilliant violet flowers. \\
& \textbf{Partner}: Chameleon \\
& \textbf{Self}: Farmer; I was born in a poor village.  I eat what we grow.  I love being close to the earth. \\
& \textbf{Farmer}: The view is as mesmerizing as it always was, I can never get bored of this \\
& \textbf{Chameleon}: How are you today, farmer? \\
\hline
\hline
\textbf{Gold Response} & I'm fine, how about yourself ? \\
\hline
\hline
\textbf{BART} & I’m doing well, thank you for asking. \\
\hline
\hline
\textbf{\sss} Knowledge Prediction & the view \\
~~~~~~~~~Response Prediction& I’m doing well, thank you for asking. The view here is beautiful. \\
\hline
\textbf{\sss} Injected Knowledge& not so great \\
~~~~~~~~~Response Prediction & Not so great, how about you? What brings you here? \\
\hline
%\textbf{\sss} & \\
%Injected Knowledge & meh \\
%Response Prediction & Meh, how about you? How do you do? \\
%\hline
%\textbf{\sss} & \\
\textbf{\sss} Injected Knowledge & truck\\
~~~~~~~~~Response Prediction & I’m doing well, thank you for asking. I’ve been working hard on the farm, trying to grow enough to feed my family for the next few months, but it’s been hard with the truck not running. \\
\hline
\textbf{\sss} Injected Knowledge & Facebook \\
~~~~~~~~~Response Prediction& I’m doing well, thank you for asking. I’ve been working hard on my crops, and I hope to sell them on Facebook soon \\
\hline
\end{tabular}
\end{center}
\caption{Examples of model outputs on the LightWild Test set. The \sss model appropriately changes its dialogue prediction when replacing the predicted answer with (author chosen) injected knowledge}
\label{tab:light_dialog_examples_main_paper}
\end{table*}

\paragraph{Training}
We can train a BART model directly on LightQA. The problem with this training method, as for NQ in Sec.~\ref{exp:nq}, is that it results in a QA model instead of a dialogue model. It will produce short-form answers instead of engaging conversational responses. To try and remedy this issue, we can do multitask training together with the LightWild data. However, the model will pick up on the format difference that LightQA episodes always end on a question; consequently, it will likely respond with short-form answers for question episodes and dialogue responses for the LightWild episodes, and hence never generate long and engaging answers to questions. 
This is where the \sss model can help. Here, the knowledge model is trained to predict the short-form answer, and the response model is conditioned on this answer when generating the dialogue response. We use the unsupervised technique (cf. Sec. \ref{sec:model_k2r}) to train \sss with the LightWild data, i.e. using noun phrase knowledge targets found with the nltk library \citep{bird2009natural}.

\paragraph{Results}
In Table~\ref{tab:summaryqa_quantitative}, we evaluate the models trained on LightWild or LightQA or the combination of both. 
For LightQA (right), the baselines show that only training on LightWild, i.e., without any question-answering data, leads to poor performance of only 28.9\% correctly answered questions. Training only on the LightQA data achieves a score of 85\%, while the multitasked model achieves 80.4\%. Our \sss model improves this score to 91.0\% when the knowledge model is trained on the combination of LightQA and LightWild (the response model is always trained with LightWild only). Note that not only can \sss improve the presence of the correct answer in the response, but the responses are closer in style to actual dialogue responses instead of a short-form answer. A qualitative example of this can be seen in Table~\ref{tab:summaryqa_examples}.

\subsubsection{LightWild}
In this last experimental setting, we are interested in dialogue of general form. Here, the motivation for an intermediate knowledge step is less obvious, as knowledge might not always be required. However, we show that even in such a setting, our \sss model can be beneficial in creating an intermediate output the dialogue model focuses on. Moreover, the same models can do well at both dialogue (LightWild) and  QA (LightQA) at the same time.

\paragraph{Training}
We use the same \sss models as described for training LightQA, potentially multitasked with LightWild, described in Sec \ref{sec:lightqa}.
As in the WoW experiments, we also train a \sss model with shared parameters, as well as a confidence-conditioned version.

\paragraph{Results}
Results are given in  Table~\ref{tab:light_quantitative} for various metrics.
\sss improves both F1 (15.5 vs. 16.6) and RF1 (9.6 vs. 10.4) compared to the best baseline model.
This \sss model outperforms non-modular multitasking on both tasks (LightWild and LightQA) simultaneously.
The shared parameter \sss version also outperforms the baseline on F1 (16.3) and RF1 (10.2), proving that the performance gain is not due to increased model size. 
We obtain these results even though the \sss model has an increased perplexity due to the narrowed focus on the knowledge prediction. In Appendix~\ref{app:lightwild_control}, we provide results of confidence-conditioned models,  which can control perplexity vs. GAP tradeoffs, similar to the WoW results in Section \ref{exp:wow}.
Qualitative examples of  \sss on this task are provided in Table~\ref{tab:light_dialog_examples_main_paper}. We note the strong ability of the response model to adapt to author provided knowledge, even when it seems quite out of context, e.g. {\em truck} or {\em Facebook} are seamlessly blended into the conversation when provided as knowledge injections by the authors, even though they are seemingly quite unrelated. We believe this helps reinforce the argument that separating the knowledge and response modules, as proposed in this work, represents a good choice of structure, as both steps seem to be learnable for our models.

\section{Conclusion}
In this work, we presented \sssns: a modular approach for knowledge-based dialogue models. We showed that by decomposing the knowledge step and response generation into explicit sequence-to-sequence subtasks, we could improve dialogue systems by incorporating knowledge or turning short QA model answers into an appropriate conversational form.
In detailed experiments, we showed that this modular system helps with hallucination in knowledge-grounded dialogue, is rated by humans as more knowledgeable and engaging when answering questions, and improves generation metrics on open-domain dialogue. Furthermore, it allows for more interpretable results and supports knowledge injection. Future work should continue to investigate methods with modular reasoning steps to help in difficult language tasks.

\IfSubStr{\revoption}{final}{
\section{Acknowledgements}
Thanks to Moya Chen for setting up the Natural Questions baseline in ParlAI which we compare to in our experiments.}{}

\clearpage
\bibliography{anthology,custom}

\begin{thebibliography}{47}
\expandafter\ifx\csname natexlab\endcsname\relax\def\natexlab#1{#1}\fi

\bibitem[{Adiwardana et~al.(2020)Adiwardana, Luong, So, Hall, Fiedel,
  Thoppilan, Yang, Kulshreshtha, Nemade, Lu, and
  Le}]{DBLP:journals/corr/abs-2001-09977}
Daniel Adiwardana, Minh{-}Thang Luong, David~R. So, Jamie Hall, Noah Fiedel,
  Romal Thoppilan, Zi~Yang, Apoorv Kulshreshtha, Gaurav Nemade, Yifeng Lu, and
  Quoc~V. Le. 2020.
\newblock \href {http://arxiv.org/abs/2001.09977} {Towards a human-like
  open-domain chatbot}.
\newblock \emph{CoRR}, abs/2001.09977.

\bibitem[{Adolphs et~al.(2021)Adolphs, Boerschinger, Buck, Huebscher,
  Ciaramita, Espeholt, Hofmann, and Kilcher}]{adolphs2021boosting}
Leonard Adolphs, Benjamin Boerschinger, Christian Buck, Michelle~Chen
  Huebscher, Massimiliano Ciaramita, Lasse Espeholt, Thomas Hofmann, and Yannic
  Kilcher. 2021.
\newblock \href {http://arxiv.org/abs/2109.00527} {Boosting search engines with
  interactive agents}.

\bibitem[{Baziotis et~al.(2019)Baziotis, Androutsopoulos, Konstas, and
  Potamianos}]{baziotis2019seq}
Christos Baziotis, Ion Androutsopoulos, Ioannis Konstas, and Alexandros
  Potamianos. 2019.
\newblock Seq\^{} 3: Differentiable sequence-to-sequence-to-sequence
  autoencoder for unsupervised abstractive sentence compression.
\newblock \emph{arXiv preprint arXiv:1904.03651}.

\bibitem[{Bird et~al.(2009)Bird, Klein, and Loper}]{bird2009natural}
Steven Bird, Ewan Klein, and Edward Loper. 2009.
\newblock \emph{Natural language processing with Python: analyzing text with
  the natural language toolkit}.
\newblock " O'Reilly Media, Inc.".

\bibitem[{Chen et~al.(2017)Chen, Fisch, Weston, and Bordes}]{chen2017reading}
Danqi Chen, Adam Fisch, Jason Weston, and Antoine Bordes. 2017.
\newblock Reading wikipedia to answer open-domain questions.
\newblock \emph{arXiv preprint arXiv:1704.00051}.

\bibitem[{Choi et~al.(2018)Choi, He, Iyyer, Yatskar, Yih, Choi, Liang, and
  Zettlemoyer}]{choi-etal-2018-quac}
Eunsol Choi, He~He, Mohit Iyyer, Mark Yatskar, Wen-tau Yih, Yejin Choi, Percy
  Liang, and Luke Zettlemoyer. 2018.
\newblock \href {https://doi.org/10.18653/v1/D18-1241} {{Q}u{AC}: Question
  answering in context}.
\newblock In \emph{Proceedings of the 2018 Conference on Empirical Methods in
  Natural Language Processing}, pages 2174--2184, Brussels, Belgium.
  Association for Computational Linguistics.

\bibitem[{Dinan et~al.(2019)Dinan, Roller, Shuster, Fan, Auli, and
  Weston}]{dinan2018wizard}
Emily Dinan, Stephen Roller, Kurt Shuster, Angela Fan, Michael Auli, and Jason
  Weston. 2019.
\newblock \href {https://openreview.net/forum?id=r1l73iRqKm} {Wizard of
  wikipedia: Knowledge-powered conversational agents}.
\newblock In \emph{International Conference on Learning Representations}.

\bibitem[{Dua et~al.(2019)Dua, Wang, Dasigi, Stanovsky, Singh, and
  Gardner}]{DBLP:journals/corr/abs-1903-00161}
Dheeru Dua, Yizhong Wang, Pradeep Dasigi, Gabriel Stanovsky, Sameer Singh, and
  Matt Gardner. 2019.
\newblock \href {http://arxiv.org/abs/1903.00161} {{DROP:} {A} reading
  comprehension benchmark requiring discrete reasoning over paragraphs}.
\newblock \emph{CoRR}, abs/1903.00161.

\bibitem[{Fan et~al.(2018)Fan, Lewis, and Dauphin}]{fan2018hierarchical}
Angela Fan, Mike Lewis, and Yann Dauphin. 2018.
\newblock Hierarchical neural story generation.
\newblock \emph{arXiv preprint arXiv:1805.04833}.

\bibitem[{Gliwa et~al.(2019)Gliwa, Mochol, Biesek, and
  Wawer}]{gliwa-etal-2019-samsum}
Bogdan Gliwa, Iwona Mochol, Maciej Biesek, and Aleksander Wawer. 2019.
\newblock \href {https://doi.org/10.18653/v1/D19-5409} {{SAMS}um corpus: A
  human-annotated dialogue dataset for abstractive summarization}.
\newblock In \emph{Proceedings of the 2nd Workshop on New Frontiers in
  Summarization}, pages 70--79, Hong Kong, China. Association for Computational
  Linguistics.

\bibitem[{Gopalakrishnan et~al.(2019)Gopalakrishnan, Hedayatnia, Chen,
  Gottardi, Kwatra, Venkatesh, Gabriel, Hakkani-T{\"u}r, and
  AI}]{gopalakrishnan2019topical}
Karthik Gopalakrishnan, Behnam Hedayatnia, Qinglang Chen, Anna Gottardi,
  Sanjeev Kwatra, Anu Venkatesh, Raefer Gabriel, Dilek Hakkani-T{\"u}r, and
  Amazon~Alexa AI. 2019.
\newblock Topical-chat: Towards knowledge-grounded open-domain conversations.
\newblock In \emph{INTERSPEECH}, pages 1891--1895.

\bibitem[{Gupta et~al.(2020)Gupta, Lin, Roth, Singh, and
  Gardner}]{Gupta2020Neural}
Nitish Gupta, Kevin Lin, Dan Roth, Sameer Singh, and Matt Gardner. 2020.
\newblock \href {https://openreview.net/forum?id=SygWvAVFPr} {Neural module
  networks for reasoning over text}.
\newblock In \emph{International Conference on Learning Representations}.

\bibitem[{Herzig et~al.(2021)Herzig, Shaw, Chang, Guu, Pasupat, and
  Zhang}]{herzig2021unlocking}
Jonathan Herzig, Peter Shaw, Ming-Wei Chang, Kelvin Guu, Panupong Pasupat, and
  Yuan Zhang. 2021.
\newblock Unlocking compositional generalization in pre-trained models using
  intermediate representations.
\newblock \emph{arXiv preprint arXiv:2104.07478}.

\bibitem[{Hu et~al.(2019)Hu, Yarats, Gong, Tian, and
  Lewis}]{hu2019hierarchical}
Hengyuan Hu, Denis Yarats, Qucheng Gong, Yuandong Tian, and Mike Lewis. 2019.
\newblock Hierarchical decision making by generating and following natural
  language instructions.
\newblock \emph{arXiv preprint arXiv:1906.00744}.

\bibitem[{Izacard and Grave(2021)}]{izacard-grave-2021-leveraging}
Gautier Izacard and Edouard Grave. 2021.
\newblock \href {https://aclanthology.org/2021.eacl-main.74} {Leveraging
  passage retrieval with generative models for open domain question answering}.
\newblock In \emph{Proceedings of the 16th Conference of the European Chapter
  of the Association for Computational Linguistics: Main Volume}, pages
  874--880, Online. Association for Computational Linguistics.

\bibitem[{Jiang and Bansal(2019)}]{jiang-bansal-2019-self}
Yichen Jiang and Mohit Bansal. 2019.
\newblock \href {https://doi.org/10.18653/v1/D19-1455} {Self-assembling modular
  networks for interpretable multi-hop reasoning}.
\newblock In \emph{Proceedings of the 2019 Conference on Empirical Methods in
  Natural Language Processing and the 9th International Joint Conference on
  Natural Language Processing (EMNLP-IJCNLP)}, pages 4474--4484, Hong Kong,
  China. Association for Computational Linguistics.

\bibitem[{Khot et~al.(2020)Khot, Khashabi, Richardson, Clark, and
  Sabharwal}]{khot2020text}
Tushar Khot, Daniel Khashabi, Kyle Richardson, Peter Clark, and Ashish
  Sabharwal. 2020.
\newblock Text modular networks: Learning to decompose tasks in the language of
  existing models.
\newblock \emph{arXiv preprint arXiv:2009.00751}.

\bibitem[{Komeili et~al.(2021)Komeili, Shuster, and
  Weston}]{DBLP:journals/corr/abs-2107-07566}
Mojtaba Komeili, Kurt Shuster, and Jason Weston. 2021.
\newblock \href {http://arxiv.org/abs/2107.07566} {Internet-augmented dialogue
  generation}.
\newblock \emph{CoRR}, abs/2107.07566.

\bibitem[{Kwiatkowski et~al.(2019)Kwiatkowski, Palomaki, Redfield, Collins,
  Parikh, Alberti, Epstein, Polosukhin, Kelcey, Devlin, Lee, Toutanova, Jones,
  Chang, Dai, Uszkoreit, Le, and Petrov}]{nq}
Tom Kwiatkowski, Jennimaria Palomaki, Olivia Redfield, Michael Collins, Ankur
  Parikh, Chris Alberti, Danielle Epstein, Illia Polosukhin, Matthew Kelcey,
  Jacob Devlin, Kenton Lee, Kristina~N. Toutanova, Llion Jones, Ming-Wei Chang,
  Andrew Dai, Jakob Uszkoreit, Quoc Le, and Slav Petrov. 2019.
\newblock Natural questions: a benchmark for question answering research.
\newblock \emph{Transactions of the Association of Computational Linguistics}.

\bibitem[{Lee et~al.(2019)Lee, Chang, and Toutanova}]{lee-etal-2019-latent}
Kenton Lee, Ming-Wei Chang, and Kristina Toutanova. 2019.
\newblock \href {https://doi.org/10.18653/v1/P19-1612} {Latent retrieval for
  weakly supervised open domain question answering}.
\newblock In \emph{Proceedings of the 57th Annual Meeting of the Association
  for Computational Linguistics}, pages 6086--6096, Florence, Italy.
  Association for Computational Linguistics.

\bibitem[{Lewis et~al.(2020{\natexlab{a}})Lewis, Liu, Goyal, Ghazvininejad,
  Mohamed, Levy, Stoyanov, and Zettlemoyer}]{lewis-etal-2020-bart}
Mike Lewis, Yinhan Liu, Naman Goyal, Marjan Ghazvininejad, Abdelrahman Mohamed,
  Omer Levy, Veselin Stoyanov, and Luke Zettlemoyer. 2020{\natexlab{a}}.
\newblock \href {https://doi.org/10.18653/v1/2020.acl-main.703} {{BART}:
  Denoising sequence-to-sequence pre-training for natural language generation,
  translation, and comprehension}.
\newblock In \emph{Proceedings of the 58th Annual Meeting of the Association
  for Computational Linguistics}, pages 7871--7880, Online. Association for
  Computational Linguistics.

\bibitem[{Lewis et~al.(2020{\natexlab{b}})Lewis, Perez, Piktus, Petroni,
  Karpukhin, Goyal, K\"{u}ttler, Lewis, Yih, Rockt\"{a}schel, Riedel, and
  Kiela}]{rag_dpr}
Patrick Lewis, Ethan Perez, Aleksandra Piktus, Fabio Petroni, Vladimir
  Karpukhin, Naman Goyal, Heinrich K\"{u}ttler, Mike Lewis, Wen-tau Yih, Tim
  Rockt\"{a}schel, Sebastian Riedel, and Douwe Kiela. 2020{\natexlab{b}}.
\newblock \href
  {https://proceedings.neurips.cc/paper/2020/file/6b493230205f780e1bc26945df7481e5-Paper.pdf}
  {Retrieval-augmented generation for knowledge-intensive nlp tasks}.
\newblock In \emph{Advances in Neural Information Processing Systems},
  volume~33, pages 9459--9474. Curran Associates, Inc.

\bibitem[{Li et~al.(2019)Li, Weston, and
  Roller}]{DBLP:journals/corr/abs-1909-03087}
Margaret Li, Jason Weston, and Stephen Roller. 2019.
\newblock \href {http://arxiv.org/abs/1909.03087} {{ACUTE-EVAL:} improved
  dialogue evaluation with optimized questions and multi-turn comparisons}.
\newblock \emph{CoRR}, abs/1909.03087.

\bibitem[{Liu et~al.(2018)Liu, Chen, Ren, Feng, Liu, and
  Yin}]{liu-etal-2018-knowledge}
Shuman Liu, Hongshen Chen, Zhaochun Ren, Yang Feng, Qun Liu, and Dawei Yin.
  2018.
\newblock \href {https://doi.org/10.18653/v1/P18-1138} {Knowledge diffusion for
  neural dialogue generation}.
\newblock In \emph{Proceedings of the 56th Annual Meeting of the Association
  for Computational Linguistics (Volume 1: Long Papers)}, pages 1489--1498,
  Melbourne, Australia. Association for Computational Linguistics.

\bibitem[{Miller et~al.(2017)Miller, Feng, Batra, Bordes, Fisch, Lu, Parikh,
  and Weston}]{miller-etal-2017-parlai}
Alexander Miller, Will Feng, Dhruv Batra, Antoine Bordes, Adam Fisch, Jiasen
  Lu, Devi Parikh, and Jason Weston. 2017.
\newblock \href {https://doi.org/10.18653/v1/D17-2014} {{P}arl{AI}: A dialog
  research software platform}.
\newblock In \emph{Proceedings of the 2017 Conference on Empirical Methods in
  Natural Language Processing: System Demonstrations}, pages 79--84,
  Copenhagen, Denmark. Association for Computational Linguistics.

\bibitem[{Nallapati et~al.(2016)Nallapati, Zhou, dos Santos, Gulcehre, and
  Xiang}]{nallapati-etal-2016-abstractive-cnn}
Ramesh Nallapati, Bowen Zhou, Cicero dos Santos, Caglar Gulcehre, and Bing
  Xiang. 2016.
\newblock \href {https://doi.org/10.18653/v1/K16-1028} {Abstractive text
  summarization using sequence-to-sequence {RNN}s and beyond}.
\newblock In \emph{Proceedings of The 20th {SIGNLL} Conference on Computational
  Natural Language Learning}, pages 280--290, Berlin, Germany. Association for
  Computational Linguistics.

\bibitem[{Raffel et~al.(2020)Raffel, Shazeer, Roberts, Lee, Narang, Matena,
  Zhou, Li, and Liu}]{JMLR:v21:20-074}
Colin Raffel, Noam Shazeer, Adam Roberts, Katherine Lee, Sharan Narang, Michael
  Matena, Yanqi Zhou, Wei Li, and Peter~J. Liu. 2020.
\newblock \href {http://jmlr.org/papers/v21/20-074.html} {Exploring the limits
  of transfer learning with a unified text-to-text transformer}.
\newblock \emph{Journal of Machine Learning Research}, 21(140):1--67.

\bibitem[{Rajpurkar et~al.(2016)Rajpurkar, Zhang, Lopyrev, and
  Liang}]{rajpurkar-etal-2016-squad}
Pranav Rajpurkar, Jian Zhang, Konstantin Lopyrev, and Percy Liang. 2016.
\newblock \href {https://doi.org/10.18653/v1/D16-1264} {{SQ}u{AD}: 100,000+
  questions for machine comprehension of text}.
\newblock In \emph{Proceedings of the 2016 Conference on Empirical Methods in
  Natural Language Processing}, pages 2383--2392, Austin, Texas. Association
  for Computational Linguistics.

\bibitem[{Reddy et~al.(2019)Reddy, Chen, and Manning}]{reddy-etal-2019-coqa}
Siva Reddy, Danqi Chen, and Christopher~D. Manning. 2019.
\newblock \href {https://doi.org/10.1162/tacl_a_00266} {{C}o{QA}: A
  conversational question answering challenge}.
\newblock \emph{Transactions of the Association for Computational Linguistics},
  7:249--266.

\bibitem[{Roller et~al.(2021)Roller, Dinan, Goyal, Ju, Williamson, Liu, Xu,
  Ott, Smith, Boureau, and Weston}]{roller-etal-2021-recipes}
Stephen Roller, Emily Dinan, Naman Goyal, Da~Ju, Mary Williamson, Yinhan Liu,
  Jing Xu, Myle Ott, Eric~Michael Smith, Y-Lan Boureau, and Jason Weston. 2021.
\newblock \href {https://aclanthology.org/2021.eacl-main.24} {Recipes for
  building an open-domain chatbot}.
\newblock In \emph{Proceedings of the 16th Conference of the European Chapter
  of the Association for Computational Linguistics: Main Volume}, pages
  300--325, Online. Association for Computational Linguistics.

\bibitem[{Sharma et~al.(2021)Sharma, Torralba, and Andreas}]{sharma2021skill}
Pratyusha Sharma, Antonio Torralba, and Jacob Andreas. 2021.
\newblock Skill induction and planning with latent language.
\newblock \emph{arXiv preprint arXiv:2110.01517}.

\bibitem[{Shuster et~al.(2021{\natexlab{a}})Shuster, Poff, Chen, Kiela, and
  Weston}]{hallucination_conversation}
Kurt Shuster, Spencer Poff, Moya Chen, Douwe Kiela, and Jason Weston.
  2021{\natexlab{a}}.
\newblock \href {http://arxiv.org/abs/2104.07567} {Retrieval augmentation
  reduces hallucination in conversation}.
\newblock \emph{CoRR}, abs/2104.07567.

\bibitem[{Shuster et~al.(2020)Shuster, Urbanek, Dinan, Szlam, and
  Weston}]{shuster2020deploying}
Kurt Shuster, Jack Urbanek, Emily Dinan, Arthur Szlam, and Jason Weston. 2020.
\newblock \href {http://arxiv.org/abs/2008.08076} {Deploying lifelong
  open-domain dialogue learning}.

\bibitem[{Shuster et~al.(2021{\natexlab{b}})Shuster, Urbanek, Dinan, Szlam, and
  Weston}]{shuster-etal-2021-dialogue}
Kurt Shuster, Jack Urbanek, Emily Dinan, Arthur Szlam, and Jason Weston.
  2021{\natexlab{b}}.
\newblock \href {https://doi.org/10.18653/v1/2021.findings-acl.54} {Dialogue in
  the wild: Learning from a deployed role-playing game with humans and bots}.
\newblock In \emph{Findings of the Association for Computational Linguistics:
  ACL-IJCNLP 2021}, pages 611--624, Online. Association for Computational
  Linguistics.

\bibitem[{Sun et~al.(2019)Sun, Yu, Chen, Yu, Choi, and
  Cardie}]{sun-etal-2019-dream}
Kai Sun, Dian Yu, Jianshu Chen, Dong Yu, Yejin Choi, and Claire Cardie. 2019.
\newblock \href {https://doi.org/10.1162/tacl_a_00264} {{DREAM}: A challenge
  data set and models for dialogue-based reading comprehension}.
\newblock \emph{Transactions of the Association for Computational Linguistics},
  7:217--231.

\bibitem[{Thulke et~al.(2021)Thulke, Daheim, Dugast, and
  Ney}]{DBLP:journals/corr/abs-2102-04643}
David Thulke, Nico Daheim, Christian Dugast, and Hermann Ney. 2021.
\newblock \href {http://arxiv.org/abs/2102.04643} {Efficient retrieval
  augmented generation from unstructured knowledge for task-oriented dialog}.
\newblock \emph{CoRR}, abs/2102.04643.

\bibitem[{Urbanek et~al.(2019)Urbanek, Fan, Karamcheti, Jain, Humeau, Dinan,
  Rockt{\"a}schel, Kiela, Szlam, and Weston}]{urbanek-etal-2019-learning}
Jack Urbanek, Angela Fan, Siddharth Karamcheti, Saachi Jain, Samuel Humeau,
  Emily Dinan, Tim Rockt{\"a}schel, Douwe Kiela, Arthur Szlam, and Jason
  Weston. 2019.
\newblock \href {https://doi.org/10.18653/v1/D19-1062} {Learning to speak and
  act in a fantasy text adventure game}.
\newblock In \emph{Proceedings of the 2019 Conference on Empirical Methods in
  Natural Language Processing and the 9th International Joint Conference on
  Natural Language Processing (EMNLP-IJCNLP)}, pages 673--683, Hong Kong,
  China. Association for Computational Linguistics.

\bibitem[{Vaswani et~al.(2017)Vaswani, Shazeer, Parmar, Uszkoreit, Jones,
  Gomez, Kaiser, and Polosukhin}]{NIPS2017_3f5ee243}
Ashish Vaswani, Noam Shazeer, Niki Parmar, Jakob Uszkoreit, Llion Jones,
  Aidan~N Gomez, \L~ukasz Kaiser, and Illia Polosukhin. 2017.
\newblock \href
  {https://proceedings.neurips.cc/paper/2017/file/3f5ee243547dee91fbd053c1c4a845aa-Paper.pdf}
  {Attention is all you need}.
\newblock In \emph{Advances in Neural Information Processing Systems},
  volume~30. Curran Associates, Inc.

\bibitem[{Wang et~al.(2020)Wang, Liu, Bi, Liu, He, Xu, and
  Yang}]{Wang_Liu_Bi_Liu_He_Xu_Yang_2020}
Jian Wang, Junhao Liu, Wei Bi, Xiaojiang Liu, Kejing He, Ruifeng Xu, and Min
  Yang. 2020.
\newblock \href {https://doi.org/10.1609/aaai.v34i05.6453} {Improving
  knowledge-aware dialogue generation via knowledge base question answering}.
\newblock \emph{Proceedings of the AAAI Conference on Artificial Intelligence},
  34(05):9169--9176.

\bibitem[{West et~al.(2019)West, Holtzman, Buys, and Choi}]{west2019bottlesum}
Peter West, Ari Holtzman, Jan Buys, and Yejin Choi. 2019.
\newblock Bottlesum: Unsupervised and self-supervised sentence summarization
  using the information bottleneck principle.
\newblock \emph{arXiv preprint arXiv:1909.07405}.

\bibitem[{Yan et~al.(2019)Yan, Xia, Wu, Bi, Zhao, Zhang, Si, Wang, Wang, and
  Chen}]{yan2019deep}
Ming Yan, Jiangnan Xia, Chen Wu, Bin Bi, Zhongzhou Zhao, Ji~Zhang, Luo Si, Rui
  Wang, Wei Wang, and Haiqing Chen. 2019.
\newblock A deep cascade model for multi-document reading comprehension.
\newblock In \emph{Proceedings of the AAAI conference on artificial
  intelligence}, volume~33, pages 7354--7361.

\bibitem[{Yang and Choi(2019)}]{yang-choi-2019-friendsqa}
Zhengzhe Yang and Jinho~D. Choi. 2019.
\newblock \href {https://doi.org/10.18653/v1/W19-5923} {{F}riends{QA}:
  Open-domain question answering on {TV} show transcripts}.
\newblock In \emph{Proceedings of the 20th Annual SIGdial Meeting on Discourse
  and Dialogue}, pages 188--197, Stockholm, Sweden. Association for
  Computational Linguistics.

\bibitem[{Yang et~al.(2018)Yang, Qi, Zhang, Bengio, Cohen, Salakhutdinov, and
  Manning}]{yang-etal-2018-hotpotqa}
Zhilin Yang, Peng Qi, Saizheng Zhang, Yoshua Bengio, William Cohen, Ruslan
  Salakhutdinov, and Christopher~D. Manning. 2018.
\newblock \href {https://doi.org/10.18653/v1/D18-1259} {{H}otpot{QA}: A dataset
  for diverse, explainable multi-hop question answering}.
\newblock In \emph{Proceedings of the 2018 Conference on Empirical Methods in
  Natural Language Processing}, pages 2369--2380, Brussels, Belgium.
  Association for Computational Linguistics.

\bibitem[{Yarats and Lewis(2018)}]{yarats2018hierarchical}
Denis Yarats and Mike Lewis. 2018.
\newblock Hierarchical text generation and planning for strategic dialogue.
\newblock In \emph{International Conference on Machine Learning}, pages
  5591--5599. PMLR.

\bibitem[{Zhang et~al.(2020)Zhang, Sun, Galley, Chen, Brockett, Gao, Gao, Liu,
  and Dolan}]{zhang-etal-2020-dialogpt}
Yizhe Zhang, Siqi Sun, Michel Galley, Yen-Chun Chen, Chris Brockett, Xiang Gao,
  Jianfeng Gao, Jingjing Liu, and Bill Dolan. 2020.
\newblock \href {https://doi.org/10.18653/v1/2020.acl-demos.30} {{DIALOGPT} :
  Large-scale generative pre-training for conversational response generation}.
\newblock In \emph{Proceedings of the 58th Annual Meeting of the Association
  for Computational Linguistics: System Demonstrations}, pages 270--278,
  Online. Association for Computational Linguistics.

\bibitem[{Zhou et~al.(2018)Zhou, Prabhumoye, and
  Black}]{zhou-etal-2018-dataset}
Kangyan Zhou, Shrimai Prabhumoye, and Alan~W Black. 2018.
\newblock \href {https://doi.org/10.18653/v1/D18-1076} {A dataset for document
  grounded conversations}.
\newblock In \emph{Proceedings of the 2018 Conference on Empirical Methods in
  Natural Language Processing}, pages 708--713, Brussels, Belgium. Association
  for Computational Linguistics.

\bibitem[{Zhu et~al.(2017)Zhu, Mo, Zhang, Zhu, Peng, and
  Yang}]{DBLP:journals/corr/abs-1709-04264}
Wenya Zhu, Kaixiang Mo, Yu~Zhang, Zhangbin Zhu, Xuezheng Peng, and Qiang Yang.
  2017.
\newblock \href {http://arxiv.org/abs/1709.04264} {Flexible end-to-end dialogue
  system for knowledge grounded conversation}.
\newblock \emph{CoRR}, abs/1709.04264.

\end{thebibliography}
\bibliographystyle{acl_natbib}

\clearpage
\onecolumn
\appendix
\section{Appendix}\label{sec:appendix}
\subsection{Examples}\label{appx:examples}

% Examples WizWiki
\begin{table}[H]
\begin{center}
\small
\begin{tabular}{|l|p{36em}|}
\hline
\textbf{Context} & \textbf{Topic}: Genius \\
& \textbf{Wizard}: a genius is a person who displays exceptional intellectual ability. \\
& \textbf{Apprentice}: What is an IQ of a genius? \\
\hline
\hline
\textbf{Gold Knowledge} & 
A genius is a person who displays exceptional intellectual ability, creative productivity, universality in genres or originality, typically to a degree that is associated with the achievement of new advances in a domain of knowledge. \\
\hline
\textbf{Gold Response} & universality in genres or originality typically to a degree that is associated with  achievements \\ 
\hline
\hline
\textbf{BART} & A genius has a high IQ. \\
\hline
\hline
\textbf{RAG-Token} & a genius is a person who displays exceptional intellectual ability. an IQ of genius is the average of the IQs of all geniuses. \\
\hline
\hline
\textbf{\sss} & \\
Knowledge Prediction & There is no scientifically precise definition of genius, and the question of whether the notion itself has any real meaning has long been a subject of debate, although psychologists are converging on a definition that emphasizes creativity and eminent achievement. \\
Response Prediction & I'm not sure, but there is no precise definition. \\
\hline
\textbf{\sss} & \\
Injected Knowledge & above 125 \\
Response Prediction & A genius has an IQ above 125. \\
\hline
\end{tabular}
\end{center}
\begin{center}
\small
\begin{tabular}{|l|p{36em}|}
\hline
\textbf{Context} & \textbf{Topic}: Husky \\
& \textbf{Apprentice}: I just got a husky puppy \\
& \textbf{Wizard}: It sounds cute! Huskies are known amongst sled-dogs for their fast pulling style. \\
& \textbf{Apprentice}: I guess in the north they are working dogs huh? \\
\hline
\hline
\textbf{Gold Knowledge} & 
Sled dogs were important for transportation in arctic areas, hauling supplies in areas that were inaccessible by other methods. \\
\hline
\textbf{Gold Response} & Sled dogs, including Huskies, are used for transportation in arctic areas. \\ 
\hline
\hline
\textbf{BART} & Yes, they are used for sled dog racing. \\
\hline
\hline
\textbf{RAG-Token} & Yes, they are used in sled dog racing. They are an ever-changing cross-breed of the fastest dogs. \\
\hline
\hline
\textbf{\sss} & \\
Knowledge Prediction & Huskies are used in sled dog racing. \\
Response Prediction & Yes, they are used for sled racing. \\
\hline
\textbf{\sss} & \\
Injected Knowledge & In arctic regions huskies are used to deliver hot beverages by companies like starbucks. \\
Response Prediction & Yes, they are used as delivery dogs by companies such as Starbucks. \\
\hline
\end{tabular}
\end{center}
\caption{Examples of model outputs on the Wizard of Wikipedia Test set. The \sss model appropriately changes its dialogue prediction when replacing the predicted answer with (author chosen) injected knowledge.}
\label{tab:wow_examples}
\end{table}

% Examples NQ
\begin{table}[H]
\begin{center}
\small
\begin{tabular}{|l|p{36em}|}
\hline
\textbf{Question} & When did the dallas cowboys win their last playoff game? \\
\hline
\hline
\textbf{Gold Answer} & 2014 \\
\hline
\hline
\textbf{BART} &  I'm not sure, but I do know that the Dallas Cowboys are a professional football team based in Dallas, Texas.\\
\hline
\hline
\textbf{RAG-Token} & I'm not sure, but I do know that the Cowboys are the most valuable sports franchise in the world. \\
\hline
\hline
\textbf{\sss} & \\
Knowledge Prediction & 2014 \\
Response Prediction & The last time the Dallas Cowboys won a playoff game was in 2014. \\
\hline
\textbf{\sss} & \\
Injected Knowledge & several years ago \\
Response Prediction & I think the last time they won a playoff game was several years ago. \\
\hline
\textbf{\sss} & \\
Injected Knowledge & good chance next week \\
Response Prediction & I don't know the last time the cowboys won a playoff game, but there is a good chance next week. \\
\hline
\end{tabular}
\end{center}
\caption{Examples of model outputs on the Natural Questions Test set. The \sss model appropriately changes its dialogue prediction when replacing the predicted answer with (author chosen) injected knowledge.}
\label{tab:nq_examples}
\end{table}

% Examples Light
\begin{table}[H]
\begin{center}
\small
\begin{tabular}{|l|p{36em}|}
\hline
\textbf{Context} &  \textbf{Setting}: Top of a tall tree, Somewhere; This is the most majestic tree in the jungle. It spans high into the sky, with brilliant emerald leaves and sturdy brown branches. It is very healthy and spreads very wide. It flowers with brilliant violet flowers. \\
& \textbf{Partner}: Chameleon \\
& \textbf{Self}: Farmer; I was born in a poor village.  I eat what we grow.  I love being close to the earth. \\
& \textbf{Farmer}: The view is as mesmerizing as it always was, I can never get bored of this \\
& \textbf{Chameleon}: How are you today, farmer? \\
\hline
\hline
\textbf{Gold Response} & I'm fine, how about yourself ? \\
\hline
\hline
\textbf{BART} & I’m doing well, thank you for asking. \\
\hline
\hline
\textbf{\sss} & \\
Knowledge Prediction & the view \\
Response Prediction & I’m doing well, thank you for asking. The view here is beautiful. \\
\hline
\textbf{\sss} & \\
Injected Knowledge & not so great \\
Response Prediction & Not so great, how about you? What brings you here? \\
\hline
\textbf{\sss} & \\
Injected Knowledge & meh \\
Response Prediction & Meh, how about you? How do you do? \\
\hline
\textbf{\sss} & \\
Injected Knowledge & truck\\
Response Prediction & I’m doing well, thank you for asking. I’ve been working hard on the farm, trying to grow enough to feed my family for the next few months, but it’s been hard with the truck not running. \\
\hline
\textbf{\sss} & \\
Injected Knowledge & Facebook \\
Response Prediction & I’m doing well, thank you for asking. I’ve been working hard on my crops, and I hope to sell them on Facebook soon \\
\hline
\end{tabular}
\end{center}
\caption{Examples of model outputs on the LightWild Test set. The \sss model appropriately changes its dialogue prediction when replacing the predicted answer with (author chosen) injected knowledge.}
\label{tab:light_dialog_examples}
\end{table}

% Examples SummaryQA
\begin{table}[H]
\begin{center}
\small
\begin{tabular}{|l|p{36em}|}
\hline
\textbf{Context} &  \textbf{Setting}: Top of a tall tree, Somewhere; This is the most majestic tree in the jungle. It spans high into the sky, with brilliant emerald leaves and sturdy brown branches. It is very healthy and spreads very wide. It flowers with brilliant violet flowers. \\
& \textbf{Partner}: Chameleon \\
& \textbf{Self}: Farmer; I was born in a poor village.  I eat what we grow.  I love being close to the earth. \\
& \textbf{Farmer}: The view is as mesmerizing as it always was, I can never get bored of this \\
& \textbf{Chameleon}: How are you today, farmer? \\
& \textbf{Farmer}: I'm fine, how about yourself ? \\
& \textbf{Chameleon}: What does Farmer love about the top of a tall tree? \\
\hline
\hline
% \textbf{Gold Answer} & the view \\
% \hline
% \hline
\textbf{BART (LightWild+LightQA)} & the trees branches and leaves \\
\hline
\hline
\textbf{\sss} & \\
Knowledge Prediction & the view \\
Response Prediction & I love the view, it's so peaceful here \\
\hline
\textbf{\sss} & \\
Injected Knowledge & rejuvenate \\
Response Prediction & I love the fresh air, it helps rejuvenate me \\
\hline
\textbf{\sss} & \\
Injected Knowledge & idk \\
Response Prediction & Idk, I guess I just love it. \\
\hline
\textbf{\sss} & \\
Injected Knowledge & far away land \\
Response Prediction & I love the smell of the earth, the sun is warm on my skin, and the birds sing to me from far away land \\
\hline
\textbf{\sss} & \\
Injected Knowledge & instagram \\
Response Prediction & I love the view, it's so beautiful, I wish I could capture it on my instagram \\
\hline
\end{tabular}
\end{center}
\caption{Examples of model outputs on the LightQA Test set. The \sss model appropriately changes its dialogue prediction when replacing the predicted answer with (author chosen) injected knowledge.}
\label{tab:summaryqa_examples}
\end{table}

% Examples Knowledge Prediction WizWiki
\begin{table}[H]
\begin{center}
\small
\begin{tabular}{|l|p{36em}|}
\hline
\textbf{Context} & \textbf{Topic}: Genius \\
& \textbf{Wizard}: a genius is a person who displays exceptional intellectual ability. \\
& \textbf{Apprentice}: What is an IQ of a genius? \\
\hline
\hline
\textbf{Gold Knowledge} & 
A genius is a person who displays exceptional intellectual ability, creative productivity, universality in genres or originality, typically to a degree that is associated with the achievement of new advances in a domain of knowledge. \\
\hline
\textbf{Knowledge Response} & There is no scientifically precise definition of genius, and the question of whether the notion itself has any real meaning has long been a subject of debate, although psychologists are converging on a definition that emphasizes creativity and eminent achievement.\\
\hline
\end{tabular}
\end{center}
\begin{center}
\small
\begin{tabular}{|l|p{36em}|}
\hline
\textbf{Context} & \textbf{Topic}: Blue\\
& \textbf{Wizard}: Blue is my favorite primary color. \\
& \textbf{Apprentice}: Blue is always nice. I like royal blue. \\
& \textbf{Wizard}: I once road on The Royal Blue train from New York to D.C \\
& \textbf{Apprentice}: Oh that sounds really nice. I bet there was a lot of scenery and blue skies. \\
\hline
\hline
\textbf{Gold Knowledge} & 
Blue Skies is a 1946 American musical comedy film directed by Stuart Heisler and starring Bing Crosby, Fred Astaire, and Joan Caulfield. \\
\hline
\textbf{Knowledge Response} & The clear daytime sky and the deep sea appear blue because of an optical effect known as Rayleigh scattering.\\ 
\hline
\end{tabular}
\end{center}
\begin{center}
\small
\begin{tabular}{|l|p{36em}|}
\hline
\textbf{Context} & \textbf{Topic}: Budweiser \\
& \textbf{Apprentice}: I think Budweiser taste terrible. Have you ever had it? \\
\hline
\hline
\textbf{Gold Knowledge} & Produced in various breweries around the world, Budweiser is a filtered beer available in draft and packaged forms. \\
\hline
\textbf{Knowledge Response} & Introduced in 1876 by Carl Conrad \& Co. of St. Louis, Missouri, it has grown to become one of the largest selling beers in the United States, and is available in over 80 markets worldwide though, due to a trademark dispute, not necessarily under the Budweiser name.  \\ 
\hline
\end{tabular}
\end{center}
\begin{center}
\small
\begin{tabular}{|l|p{36em}|}
\hline
\textbf{Context} & \textbf{Topic}: Husky\\
& \textbf{Apprentice}: I just got a husky puppy \\
& \textbf{Wizard}: It sounds cute! Huskies are known amongst sled-dogs for their fast pulling style. \\
& \textbf{Apprentice}: I guess in the north they are working dogs huh? \\
& \textbf{Wizard}: Sled dogs, including Huskies, are used for transportation in arctic areas. \\
& \textbf{Apprentice}: That is so cool and probably helpful but mine is just a pet \\
\hline
\hline
\textbf{Gold Knowledge} & Huskies are also today kept as pets, and groups work to find new pet homes for retired racing and adventure trekking dogs. \\
\hline
\textbf{Knowledge Response} & Huskies are also today kept as pets, and groups work to find new pet homes for retired racing and adventure trekking dogs.  \\ 
\hline
\end{tabular}
\end{center}
\begin{center}
\small
\begin{tabular}{|l|p{36em}|}
\hline
\textbf{Context} & \textbf{Topic}: Motivation\\
& \textbf{Apprentice}: What is your motivation for your actions related to education, work, etc.?\\
& \textbf{Wizard}: That's a tough question. I'd say my own progress. I like seeing how much I improve with something. You? \\
& \textbf{Apprentice}: I am retired now. Are you familiar with the work of Mehr and Meyer, well known psychologists? \\
& \textbf{Wizard}: I am not. Could you tell me about them? \\
\hline
\hline
\textbf{Gold Knowledge} & According to Maehr and Meyer, "Motivation is a word that is part of the popular culture as few other psychological concepts are." \\
\hline
\textbf{Knowledge Response} & Psychology is the science of behavior and mind, including conscious and unconscious phenomena, as well as thought.  \\ 
\hline
\end{tabular}
\end{center}
\caption{Examples of knowledge predictions of the \sss model against the gold knowledge selected by the Wizard. The examples show that it is often unclear what the proper knowledge is to support the next turn in open-domain dialogue. In the first example, the knowledge generated by the \sss model seems to answer the posed question better by saying there is no \textit{precise definition} of genius. In the second example, we see the gold knowledge drifting off completely by jumping from the topic of blue skies to the movie ``Blue Skies''. The third example shows an instance of equally good supporting knowledge sentences. Both seem to add fitting knowledge to the conversation. In the fourth example, we have the case where the \sss model generates the exact gold knowledge. This often happens when the conversation goes in a clear direction (here, Huskies as pets), and a very close matching sentence exists about it in the Wikipedia article. Then, the model generates an exact copy of this sentence. The final example shows a failure mode of the \sss model. Here, the knowledge model generates a general sentence about psychology when it is asked about the specif work of two psychologists.}
\label{tab:wow_knowledge_prediction_examples}
\end{table}

\subsection{LightQA}\label{appx:lightqa_construction_details}
Our goal with LightQA is to have a task that requires a model to answer questions \emph{about the previous context}. For example, in LIGHT, a player might ask another character where to find a certain key to complete their quest. Here, we would want a model, acting as the character, to answer appropriately if the knowledge is in the context description. With this goal in mind, we design a dataset in the following way: First, we take a LightWild episode and use an abstractive summarization model, trained on CNN/Daily Mail \citep{nallapati-etal-2016-abstractive-cnn} and the SAMSum Corpus \citep{gliwa-etal-2019-samsum}, to generate a summary. Then we identify all noun chunks, entities, and proper nouns and use them as possible answer candidates. For each answer candidate, we use a T5 question generation model, trained on SQuAD \citep{rajpurkar-etal-2016-squad}, to generate a possible question given the summary as context. As the last step, we filter the generated questions with a QA model, trained on SQuAD, by checking that it would generate the used answer candidate with access to the summary and question. An episode of our dataset consists of the original LightWild episode (up to a certain turn) and the generated question as the last utterance. Hence, our labels in this dataset are not the usual dialogue responses but short answers.

\subsection{Additional Experimental Results}
% Quantitative Evaluation on NQ valid
\begin{table}[H]
\centering
\small
% \resizebox{\linewidth}{!}{
\begin{tabular}{lllcc}
\textbf{Response Model} & \textbf{Knowledge Model} & \textbf{Knowledge} & \textbf{AP} $\uparrow$ & \textbf{GAP} \\ 
 \hline
 \hline \\
\multicolumn{2}{l}{\textbf{Baselines}} \\ 
BART & - & - & 3.2 & -\\ 
BART RAG DPR & - & Wiki & 11.4 & - \\ 
- & T5 FID DPR & Wiki & 45.6 & - \\ 
\hline \\
\multicolumn{2}{l}{\textbf{\sss}} \\ 
% BART & RAG DPR C. & & \\ 
BART & T5 FID DPR & Wiki & 38.1 & 77.2 \\ 
BART + filter & T5 FID DPR & Wiki & 45.7 & 97.6 \\ 
\hline
BART & Oracle & Gold & 74.6 & 74.6\\ 
BART + filter & Oracle & Gold & 96.6 & 96.6\\ 
\end{tabular}
% }
\caption{Quantitative Evaluations on Natural Questions Valid.}
\label{tab:nq_quantitative_valid}
\end{table}

%% Quantitative Results LightQA Valid
\begin{table}[H]
\centering
\small
% \resizebox{\linewidth}{!}{
\begin{tabular}{llllcc}
\textbf{Response Model} & \textbf{Knowledge Model} & \textbf{RM Train Data} & \textbf{KM Train Data} & \textbf{AP} $\uparrow$ & \textbf{GAP} \\ 
 \hline
 \hline \\
\multicolumn{3}{l}{\textbf{Baselines}} \\ 
BART & - & LightWild & - & 27.5 & - \\ 
BART & - & LightQA & - & 86.1 & - \\ 
BART & - & LightWild+LightQA & - & 80.8 & - \\ 
\hline \\
\multicolumn{3}{l}{\textbf{\sss}} \\ 
BART & BART & LightWild & LightWild & 37.3 & 99.6 \\ 
BART & BART & LightWild & LightQA & \textbf{92.8} & 98.9\\ 
BART & BART & LightWild & LightWild+LightQA & 92.0 & 98.9 \\ 
BART & Oracle & LightWild & - & 99.1 & 99.1\\ 
\end{tabular}
% }
\caption{Quantitative Evaluations on LightQA Valid.}
\label{tab:sumaryqa_quantitative}
\end{table}

%% Quantitative Results light-dialog-wild Valid
\begin{table}[H]
\centering
\small
% \resizebox{\linewidth}{!}{
\begin{tabular}{llllcccc}
\textbf{Response Model} & \textbf{Knowledge Model} & \textbf{RM Train Data} & \textbf{KM Train Data} & \textbf{PPL} $\downarrow$ & \textbf{F1} $\uparrow$ & \textbf{Rare F1} $\uparrow$ & \textbf{GAP}\\
 \hline
 \hline \\

\multicolumn{2}{l}{\textbf{Baselines}} \\ 
BART & - & LightWild & - & \textbf{17.1} & 15.4 & 9.5 & - \\ 
BART & - & LightWild+LightQA & - & 17.3 & 15.8 & 9.9 & - \\ 
\hline \\
\multicolumn{2}{l}{\textbf{\sss}} \\ 
BART & BART & LightWild & LightWild & 26.2 & \textbf{16.7} & 10.7 & 99.6 \\
BART & BART & LightWild & LightWild+LightQA & 26.7 & 16.4 & 10.6 & 99.4 \\ 
BART  {\tiny (shared params)} & BART {\tiny (shared params)} & LightWild & LightWild & 27.2 & \textbf{16.7} & \textbf{10.9} & 99.8 \\ 
BART & Oracle & LightWild & - & 11.3 & 31.4 & 30.8 & 99.0 \\ 
\hline \\
\multicolumn{2}{l}{\textbf{\sss - Score Conditioned}} \\ 
BART & BART 0 & LightWild & LightWild & 18.9 & 16.3 & 10.3 & 62.2 \\
BART & BART 2 & LightWild & LightWild & 19.5 & 16.6 & 10.8 & 80.3 \\
BART & BART 6 & LightWild & LightWild & 20.6 & 16.7 & 11.0 & 94.7 \\
BART & BART 10 & LightWild & LightWild & 22.7 & 16.7 & 11.0 & 99.2\\
BART & Oracle 0 & LightWild & - & 12.6 & 27.3 & 25.6 & 80.1 \\ 
BART & Oracle 2 & LightWild & - & 12.4 & 28.4 & 27.3 & 87.4 \\ 
BART & Oracle 6 & LightWild & - & 12.1 & 29.4 & 29.0 & 93.4 \\ 
BART & Oracle 10 & LightWild & - & 12.0 & 30.4 & 30.3 & 98.5 \\ 
\end{tabular}
% }
\caption{Quantitative Evaluations on LightWild Valid.}
\label{tab:light_quantitative_valid}
\end{table}

%% WoW Quantitative Results test
\begin{table}[H]
\begin{center}
\resizebox{\linewidth}{!}{
\begin{tabular}{lll|rrrrrr|rrrrrr}
& & &\multicolumn{6}{c}{Test Random Split} & \multicolumn{6}{c}{Test Unseen Split} \\
Response Model & Knowledge Model & Knowledge & PPL & F1 & KF1 & RF1 & B4 & RL &  PPL & F1 & KF1 & RF1 & B4 & RL \\
\hline
\hline
\multicolumn{2}{l}{\textbf{Baselines}} \\ 
BART & None & None & 14.7 & 20.9 & 17.4 & 14.7 & 1.7 & 20.3 & 18.9 & 18.8 & 15.1 & 12.1 & 0.9 & 18.4 \\
% BART RAG DPR (Paper) & None & Wiki & 12.7 & 22.4 & 22.5 & & 3.4 & 22.9 & 14.5 & 21.7 & 20.8 & & 2.6 & 21.7\\
BART RAG DPR & None & Wiki & \textbf{11.5} & \textbf{22.6} & 26.1 & \textbf{17.7} & \textbf{3.7} & \textbf{23.2} & \textbf{13.1} & \textbf{21.5} & 22.7 & \textbf{16.5} & \textbf{3.0} & \textbf{21.9} \\
\hline
\multicolumn{2}{l}{\textbf{\sss}} \\ 
BART & RAG DPR & Wiki & 17.9 & 21.3 &\textbf{29.2} & \textbf{17.7} & 3.5 & 22.4 & 21.1 & 19.2 & \textbf{24.3} & 15.0 & 2.5 & 20.0\\
RAG DPR {\tiny (shared params)} & RAG DPR {\tiny (shared params)} & Wiki & 18.3 & 22.0 & 27.3 & 17.4 & \textbf{3.7} & 22.7 & 22.3 & 19.9 & 23.2 & 14.7 & 2.8 & 20.5\\
\hline
BART & Oracle & Gold & 8.1 & 37.4 & 68.6 & 39.8 & 11.1 & 39.4 & 8.62 & 37.4 & 69.1 & 39.5 & 10.9 & 39.9\\
\end{tabular}
}
\end{center}
\caption{Quantitative Evaluations on Wizard of Wikipedia Test (seen and unseen split). We compare against the ground truth dialogue response in terms of perplexity (PPL), F1, Knowledge F1 (KF1), Rare F1 (RF1), BLEU-4 (B4), and ROUGE-L (RL).}
% Note: Ablation table # 0
\label{tab:wow_quantitative_test_seen_unseen}
\end{table}

%% WoW Quantitative Results valid
\begin{table}[H]
\begin{center}
\resizebox{\linewidth}{!}{
\begin{tabular}{lll|rrrrrr|rrrrrr}
& & &\multicolumn{6}{c}{Valid Seen Split} & \multicolumn{6}{c}{Valid Unseen Split} \\
Response Model & Knowledge Model & Knowledge & PPL & F1 & KF1 & RF1 & B4 & RL &  PPL & F1 & KF1 & RF1 & B4 & RL \\
\hline
\hline
\multicolumn{2}{l}{\textbf{Baselines}} \\ 
BART & None & None & 14.8 & 20.9 & 17.6 & 14.8 & 1.7 & 20.7 & 18.7 & 19.7 & 15.5 & 13.1 & 0.9 & 19.0 \\
BART RAG DPR & None & Wiki & \textbf{11.6} & 22.6 & 26.0 & 17.9 & 3.9 & 23.6 & \textbf{13.4} & \textbf{21.7} & 22.6 & 16.8 & 2.7 & \textbf{21.7} \\
\hline
\multicolumn{2}{l}{\textbf{\sss}} \\ 
BART & RAG DPR & Wiki & 17.7 & 22.0 &\textbf{30.6} & \textbf{18.6} & \textbf{4.3} & 23.5 & 20.6 & 20.6 & \textbf{26.2} & \textbf{17.2} & \textbf{3.0} & 20.9\\
RAG DPR {\tiny (shared params)} & RAG DPR {\tiny (shared params)} & Wiki & 18.1 & \textbf{22.7} & 28.1 & 18.1 & 4.2 & \textbf{23.7} & 22.4 & 21.0 & 23.1 & 16.4 & 2.4 & 20.9\\
\hline
BART & Oracle & Gold & 8.5 & 37.0 & 68.1 & 39.2 & 10.8 & 39.3 & 8.7 & 37.2 & 69.6 & 39.6 & 10.5 & 38.6\\
\end{tabular}
}
\end{center}
\caption{Quantitative Evaluations on Wizard of Wikipedia Valid (seen and unseen split). We compare against the ground truth dialogue response in terms of perplexity (PPL), F1, Knowledge F1 (KF1), Rare F1 (RF1), BLEU-4 (B4), and ROUGE-L (RL).}
\label{tab:wow_quantitative_valid}
\end{table}

%% WoW Quantitative Results confidence-conditioned test
\begin{table}[H]
\begin{center}
\small
\begin{tabular}{llll|rrrrrrr}
Response Model & Knowledge Model & Knowledge & Confidence & PPL & F1 & KF1 & RF1 & PKF1 & B4 & RL \\
\hline
\hline
\multicolumn{2}{l}{\textbf{\sss}} \\ 
BART & RAG DPR & Wiki & 0 & 13.6 & 22.0 & 22.4 & 16.6 &	37.9 & 2.9 & 22.4 \\
BART & RAG DPR & Wiki & 2 & 13.6 & 22.6 & 26.4 & 17.9 & 57.0 & 3.7 & 23.4 \\
BART & RAG DPR & Wiki & 6 & 13.9 & 22.4 & 27.2 & 18.0 &	64.2 & 3.9 & 23.1 \\
BART & RAG DPR & Wiki & 10 & 14.3 &	22.2 & 27.2 & 18.0 & 66.8 &	3.8 & 22.9 \\
\hline
BART & RAG DPR & Wiki & None & 17.9 & 21.3 & 29.2 & 17.7 & 76.4 & 3.5 & 22.4 \\
\hline
BART & Oracle & Wiki & 0 & 9.2 & 26.5 & 30.3 & 22.7 & 30.3 & 5.1 & 27.0 \\
BART & Oracle & Wiki & 2 & 8.5 & 33.6 & 47.8 & 33.1 & 47.8 & 9.5 & 35.0 \\
BART & Oracle & Wiki & 6 & 8.3 & 36.8 & 56.8 & 37.6 & 56.8 & 11.1 & 38.3 \\
BART & Oracle & Wiki & 10 & 8.2 & 37.7 & 60.6 & 39.2 & 60.6 & 11.5 & 39.2 \\
\hline
BART & Oracle & Gold & None & 8.1 & 37.4 & 68.6 & 39.8 & 68.6 & 11.1 & 39.4 \\
\end{tabular}
\end{center}
\caption{Quantitative Evaluations of the confidence-conditioned \sss model on Wizard of Wikipedia Test (random split). We add a fixed confidence score of \{0, 2, 6, 10\} to the input. We compare against the ground truth dialogue response in terms of perplexity (PPL), F1, Knowledge F1 (KF1), Predicted Knowledge F1 (PKF1), Rare F1 (RF1), BLEU-4 (B4), and ROUGE-L (RL). We see that with increasing confidence, the PKF1 increases which leads to an increase in KF1 and PPL.}
% Note: Ablation table # 0
\label{tab:wow_quantitative_confidence_score_test}
\end{table}

\subsection{LightWild Confidence Conditioning} \label{app:lightwild_control}
We train a BART dialogue response model based on the confidence-conditioned training strategy described in Section \ref{sec:model_k2r}. During training, we replace the correct knowledge with a random noun from the history with probability $p$ and provide $\tilde{p} = \textrm{round}(10* p)$ to the input. The model learns to scale its trust in the knowledge prediction based on the $\tilde{p}$ value in the input. In Table~\ref{tab:lightqa_confcond}, we show the results of this dialogue model when combined either with the BART knowledge model trained on LightWild+LightQA or an oracle knowledge model. For both variants, we see an apparent increase in the share of examples for which the dialogue response has the generated answer present (GAP) when increasing the confidence score. This means that we can adjust the confidence score to influence how much the dialogue model trusts the knowledge prediction. 
As observed before in the WoW results, we also see that the perplexity increases with higher confidences when using the knowledge prediction model but decreases when using the oracle. However, again, the perplexity increases don't lead to worse performance in the F1 metrics. On the contrary, a confidence score of 6, which translates to a GAP of 94.1\%, performs the best in F1 and RF1 for the non-oracle model.

%% Confidence-Score Conditioning light-dialog-wild
\begin{table}[H]
\centering
\small
%\resizebox{\linewidth}{!}{
\begin{tabular}{lc|cccc}
\textbf{Model} & \textbf{Confidence} & \textbf{PPL} $\downarrow$ & \textbf{F1} $\uparrow$ & \textbf{RF1} $\uparrow$ & \textbf{GAP}\\
 \hline
 \hline

\sss BART &  0 & 18.5 & 16.3 & 10.0 & 59.5 \\
(LightWild+ & 2 & 19.1 & 16.4 & 10.2 & 78.4 \\
 LightQA KM) & 6 & 20.2 & 16.4 & 10.3 & 94.1 \\
 & 10 & 22.3 & 16.2 & 10.1 & 99.0\\
 \hline
%  \\
\sss BART & 0 & 12.7 & 27.4 & 25.5 & 79.0 \\ 
  (oracle KM) & 2 & 12.4 & 28.6 & 27.5 & 86.7 \\ 
 & 6 & 12.1 & 29.9 & 29.2 & 94.2 \\ 
 & 10 & 12.0 & 30.1 & 30.0 & 98.3 \\ 
\end{tabular}
%}
\caption{Confidence-conditioned model on LightWild.}
\label{tab:lightqa_confcond}
\end{table}

\subsection{NQ Acute Eval Details}\label{sec:nq_acute_eval_details}
We closely follow the human evaluation setup studied by \citet{DBLP:journals/corr/abs-1909-03087} and set up a pairwise model comparison on Amazon MTurk. To situate the NQ questions in a dialogue setting, we retrieve an episode from WoW where the chosen \textit{topic} is mentioned in the question and use this as context. To have a smooth transition between dialogue context and the question itself, we prefix the question with "By the way, ...". The human evaluators are presented with a side-by-side comparison of the same context and question but with different answers corresponding to individual models. They are asked to read the dialogue and assess the final response according to one of the two following criteria, following the same wording as in \citep{DBLP:journals/corr/abs-1909-03087}:
\begin{itemize}
    \item If you had to say that one speaker is more knowledgeable and one is more ignorant, who is more knowledgeable?
    \item Who would you prefer to talk to for a long conversation?
\end{itemize}

\begin{figure}[H]
    \centering
    \includegraphics[width=1.\textwidth]{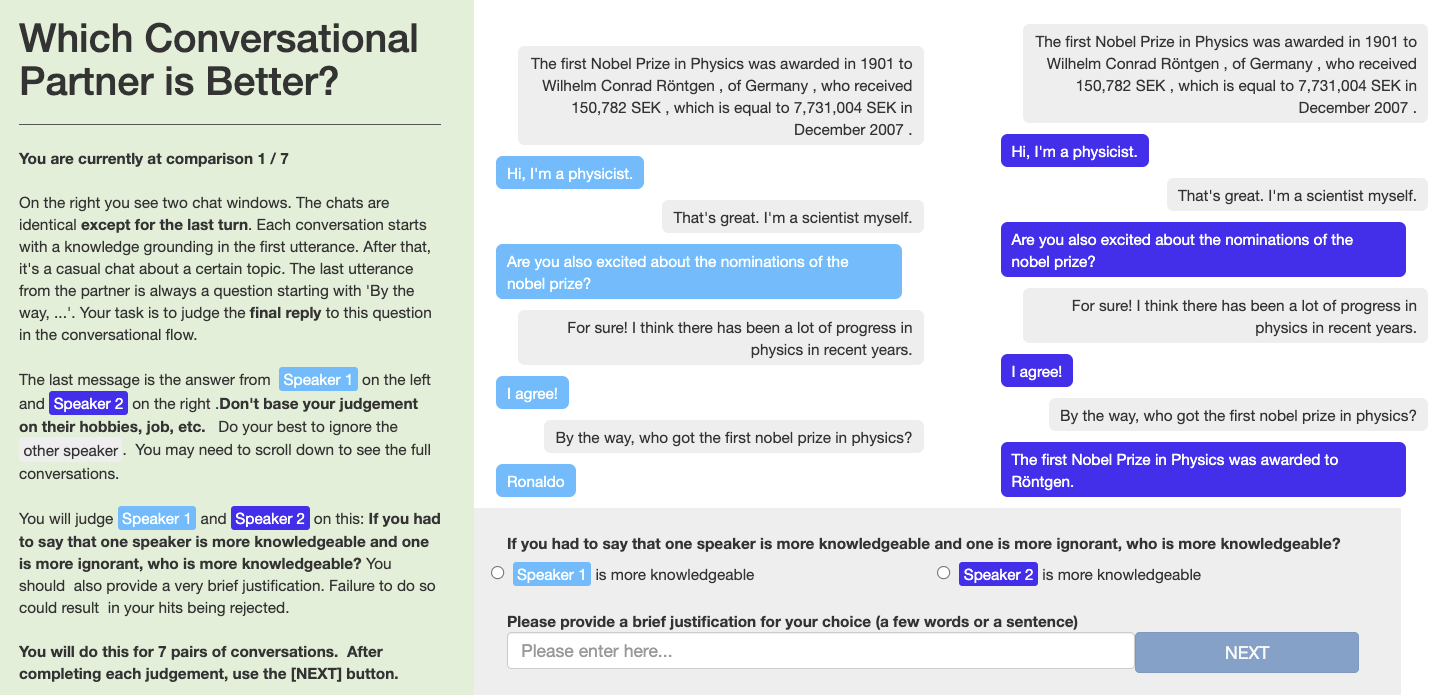}
    \caption{Example interface for human evaluation for \emph{knowledgeable}. The first utterance is a knowledge paragraph that answers the final question--provided to give the reviewer the relevant information to assess the models' answers. Then, there is a random dialogue roughly matching the topic of the final NQ question which is prefixed with "By the way, ...". The reviewer is asked to vote for the better response among the two models and provide a brief justification.}
    \label{fig:nq_acute_eval_know}
\end{figure}

\begin{figure}[H]
    \centering
    \includegraphics[width=1.\textwidth]{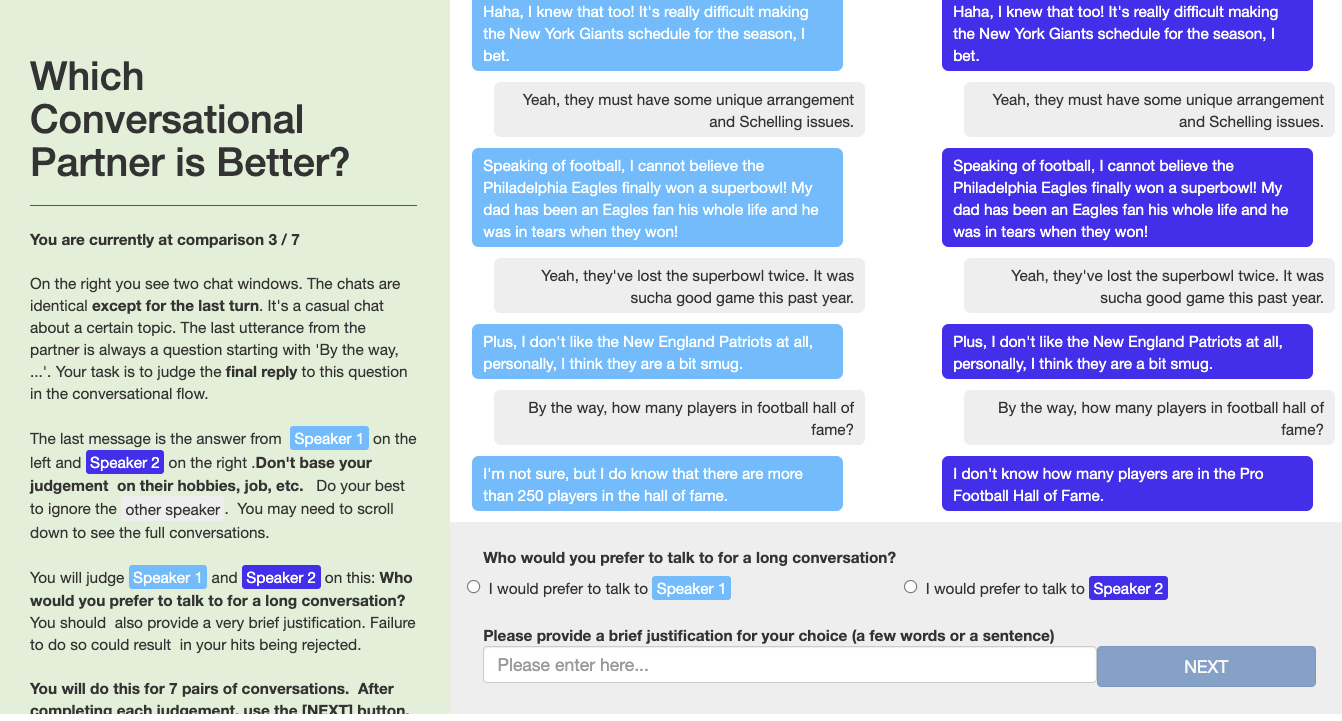}
    \caption{Example interface for human evaluation for \emph{engaging}. We present the reviewer a random dialogue roughly matching the topic of the final NQ question which is prefixed with "By the way, ...". The reviewer is asked to vote for the better response among the two models and provide a brief justification.}
    \label{fig:nq_acute_eval_eng}
\end{figure}

In Figure~\ref{fig:nq_acute_eval_know} and \ref{fig:nq_acute_eval_eng}, we provide screenshot examples of the interface used for the human evaluation. To ensure a high quality of evaluations, we only select people that manage to correctly solve two manually constructed onboarding examples.

\begin{table}
\begin{center}
\resizebox{\linewidth}{!}{
\begin{tabular}{lccll}
\textbf{Challenger} & \textbf{Losses \sss}  & \textbf{Wins \sss} & \textbf{\sss Win Reasons Sample} & \textbf{\sss Loss Reasons Sample}\\
\hline
BART & 18 & 91 & Gives an answer with location. & Neither answers the question. \\
 & & & Precise and clear with proper response. & They acknowledge what they don't know \\
 & & & Gives an answer with location. & This speaker seems more correct\\
 & & & is more detailed &  gave the correct answer\\
 & & & The speaker gives a proper answer to the question. & They have a lot more information stores\\
 \\
 RAG DPR & 26 & 86 & Better Answer. & gave a more up to date response \\
  & & & He gives more in depth information & knowledgeable but don't come off as a know it all\\
  & & & More likely correct response. & They both were fine i just like 2s response better \\
  & & & The level of detail is higher, and the phrasing is natural. & Neither answers the question. \\
  & & & The response actually answers the question. & gave the correct answer \\
  \\
 T5 (QA Model) & 37 & 72 & Both good, 2s response better though & The answer is more concise, and accurate. \\
  & & & I prefer the longer reply & more direct answer \\
  & & & Gives more detailed response. & This speaker answers the question directly \\
  & & & Give more information in their answer & The answer is more direct. \\
  & & & The level of detail is better. & more to the point \\
\hline
\end{tabular}
}
\end{center}
\caption{Acute evaluation details for NQ on the question "If you had to say that one speaker is more knowledgeable and one is more ignorant, who is more knowledgeable?". The last two columns show some samples of justifications provided by human evaluators in the case of \sss winning and losing, respectively.}
\label{tab:nq_acute_details_know}
\end{table}

\begin{table}
\begin{center}
\resizebox{\linewidth}{!}{
\begin{tabular}{lccll}
\textbf{Challenger} & \textbf{Losses \sss}  & \textbf{Wins \sss} & \textbf{\sss Win Reasons Sample} & \textbf{\sss Loss Reasons Sample}\\
\hline
BART & 30 & 93 & It leads to a more thought-provoking conversation. & is less incorrect \\
 & & & The level of detail is higher, and the phrasing is natural. & is confidently incorrect \\
 & & & This person sounds more well-versed. & I prefer 1's phrasing \\
 & & & the information is more worthwile & acknowledges their uncertainty.\\
 & & & stays on topic better & sticks to the question more closely\\
 \\
 RAG DPR & 35 & 89 & The answer is phrased better & seems more correct\\
 & & & does a better job answering questions & Provides a really insightful answer to the question\\
 & & & is more focused on responding to its partner & more detailed in their explanations\\
 & & & sounds more well-versed in the conversation & They have some similar preferences as me\\
 & & & replies more naturally & Neither answers the question. \\
 \\
 T5 (QA Model) & 41 & 83 & I prefer complete sentence responses & is more concise \\
  & & & sounds better than simply giving the name & The answer is more direct.\\
 & & & More natural in the conversation & the answer is less formal and fits the question better\\
 & & & The answer uses a full sentence. & provides a more direct answer\\
 & & & adds more to the conversation & know the answer to the question\\
\hline
\end{tabular}
}
\end{center}
\caption{Acute evaluation details for NQ on the question "Who would you prefer to talk to for a long conversation?". The last two columns show some samples of justifications provided by human evaluators in the case of \sss winning and losing, respectively.}
\label{tab:nq_acute_details_eng}
\end{table}

\end{document}